%% file: main.tex
\documentclass{article}
\usepackage[accepted]{icml2018}

\usepackage{times}
\usepackage{graphicx} 
\usepackage{subfigure} 

\usepackage{natbib}

\usepackage{algorithm}
\usepackage{algorithmic}

\usepackage{format}

\usepackage{hyperref}
\usepackage{amsmath} 
\usepackage{enumerate}
\usepackage{dsfont}
\usepackage{graphicx} 
\usepackage{xcolor}
\usepackage{ mathrsfs }

\newcommand{\qo}{q}

\newcommand{\K}{\vv k}

\newcommand{\MB}{\mathcal{N}}

\icmltitlerunning{Stein Variational Message Passing for Continuous Graphical Models}

\begin{document}

\twocolumn[
\icmltitle{Stein Variational Message Passing for Continuous Graphical Models}

\icmlsetsymbol{equal}{*}

\begin{icmlauthorlist}
\icmlauthor{Dilin Wang*}{austin}
\icmlauthor{Zhe Zeng*}{zju}
\icmlauthor{Qiang Liu}{austin}
\end{icmlauthorlist}

\icmlaffiliation{austin}{Department of Computer Science, The University of Texas at Austin}
\icmlaffiliation{zju}{School of Mathematical Sciences, Zhejiang University}

\icmlcorrespondingauthor{Dilin Wang}{dilin@cs.utexas.edu}
\icmlcorrespondingauthor{Qiang Liu}{lqiang@cs.utexas.edu}
\vskip 0.3in
]

\printAffiliationsAndNotice{\icmlEqualContribution} 

\begin{abstract}
We propose a novel distributed inference algorithm for continuous graphical models, 
by extending Stein variational gradient descent (SVGD) \citep{liu2016stein} to leverage the Markov dependency structure of the distribution of interest.  
Our approach combines SVGD with a set of structured local kernel functions
defined on the Markov blanket of each node, 
which alleviates the curse of high dimensionality and 
simultaneously yields a distributed algorithm for decentralized inference tasks. 
We justify our method with theoretical analysis and show that 
the use of local kernels can be viewed as a new type of \emph{localized} approximation that matches the target distribution 
on the conditional distributions of each node over its Markov blanket. 
Our empirical results show that our 
method outperforms a variety of baselines including standard MCMC and particle message passing methods. 
\end{abstract}

\input{tex/introduction}


\input{tex/background}

\input{tex/theory}

\input{tex/theory_graphical}
%
\input{tex/experiments}

\section{Conclusion}
In this paper,
we propose a particle-based distributed inference algorithm 
for approximate inference on continuous graphical models based on Stein
variational gradient descent (SVGD).
Our approach leverages the inherent graphical structures to improve the performance in high dimensions, 
and also incorporates the key advantages of gradient optimization compared to traditional PMP methods. 


\section*{Acknowledgements} 
This work is supported in part by NSF CRII 1565796. 
\vspace{-8pt}

\bibliography{svgd}
\bibliographystyle{icml2018}

\newpage\clearpage
\input{tex/appendix}

\end{document}

%% file: tex/introduction.tex
\section{Introduction}

Probabilistic graphical models, such as Markov random fields (MRFs) and Bayesian networks, 
provide a powerful framework for representing complex stochastic dependency structures between a large number of random effects \citep{pearl1988probabilistic, lauritzen1996graphical}. 
A key challenge, however, is to develop computationally efficient inference algorithms to approximate important integral quantities related to distributions of interest.  
Variational message passing methods, notably belief propagation \citep{pearl1988probabilistic, yedidia2003understanding},
provide one of the most powerful frameworks for approximate inference in graphical models \citep{Wainwright08}. 
In addition to accurate approximation, 
message passing algorithms perform inference in a distributed fashion by passing messages between variable nodes to propagate uncertainty, well suited for 
\emph{decentralized inference} tasks such as sensor network localization \citep[e.g.,][]{ihler2005nonparametric}.  

Unfortunately, standard belief propagation (BP)  
is best applicable only to discrete or Gaussian variable models. 
Significant additional challenges arise when applying BP to continuous, non-Gaussian graphical models, 
 because the exact BP updates involve intractable integration.
 As a result, the existing continuous variants of BP require additional particles or non-parametric approximation \citep[e.g.,][]{ihler2005nonparametric, ihler2009particle, sudderth2010nonparametric, song2011kernel}, 
 which deteriorates the accuracy and stability. 
 Another key aspect that was not well discussed in the literature is that the traditional continuous BP methods are gradient-free, in that they do not use the gradient of the density function. 
Although this makes the methods widely applicable, 
their performance can be significantly improved by 
incorporating the gradient information whenever available.

Stein variational gradient descent (SVGD) \citep{liu2016stein}
is a recent particle-based variational inference algorithm
that combines the advantages of variational inference and particle-based methods, and efficiently leverages the gradient information for continuous inference. 
Unlike traditional variational inference that constructs parametric approximation of the target distribution by minimizing KL divergence, 
SVGD directly approximates the target distribution with a set of particles, 
which is iteratively updated following a velocity field that decreases the KL divergence 
with the fastest speed among all possible velocity fields in a reproducing kernel Hilbert space (RKHS) of a positive definite kernel. 
This makes SVGD inherit the theoretical consistency of particle methods \citep{liu2017stein}, 
while obtaining the fast practical convergence thanks to its deterministic updates. 
The goal of this work is to adapt SVGD for distributed inference of continuous graphical models. 

In principle, one can directly apply SVGD to continuous graphical models. 
However, 
standard SVGD does not yield a distributed message passing like BP, 
because its update involves a kernel function, which is defined on all the variable dimensions, introducing additional dependency beyond the Markov blanket of the graphical model of interest.
In addition, the use of the global kernel function on all the variables  
also deteriorates the performance in high dimensions; 
our empirical findings show that although SVGD tends to perform exceptionally well in estimating the mean parameters, it becomes less sample efficient in terms of estimating the variances as the dimension increases. 

In this paper, we improve SVGD 
to take advantage of the inherent dependency structure of graphical models for better distributed inference.
Instead of using a global kernel function,
we associate each node with a local kernel function that only depends on the Markov blanket of each node. 
This simple modification allows us to turn SVGD into a distributed message passing algorithm, 
and simultaneously alleviates the curse of dimensionality. 

Theoretically, our method extends the original SVGD in two significant ways: 
\emph{i}) it uses different kernel functions for different coordinates (or variable nodes),  
 justified with a theoretical analysis that extends the results of \citet{liu2016kernelized, liu2016stein}; 
\emph{ii}) it uses a local kernel over the Markov blanket for each node, 
which, compared with the typical global kernel function,  
can be viewed as introducing a type of deterministic approximation to trade for better sample efficiency. 
Our empirical results show that our method outperforms a variety of baseline methods, including the typical SVGD and Monte Carlo, and particle message passing (PMP). 
%
We note that 
a similar idea is independently and concurrently proposed by 
\citet{zhuo2018message} in the same conference proceeding.  

%% file: tex/background.tex
\newcommand\eqdef{\stackrel{\mathclap{\scriptsize\mbox{def}}}{=}}
\section{Background}

We introduce the background of 
Stein variational gradient descent (SVGD) and  
kernelized Stein discrepancy (KSD),  
which forms the foundation of our work.
For notation, we denote by $x = [x_1, \ldots, x_d]$ a vector in $\RR^d$ 
and $\{x^\ell\}_{\ell=1}^n$ a set of $n$ vectors. 

\paragraph{Stein Variational Gradient Descent (SVGD)}
Let $p(x)$ be a positive differentiable density function on $\RR^d$.  
Our goal is to find a set of points (or particles) $\{x^\ell\}_{\ell=1}^n$ 
to approximate $p$ so that $\sum_{\ell=1}^n f(x^\ell)/n \approx \E_p[f]$ for general test functions $f$. 
SVGD achieves this by iteratively updating the particles with deterministic transforms of form 
$$
x^\ell \leftarrow x^\ell + \epsilon \ff(x^\ell), ~~~~~~~~~\forall \ell = 1, \ldots, n,
$$
where $\epsilon$ is a small step size and $\ff$ is a velocity field that decides the update directions of the particles. 
The key problem is to choose an optimal velocity field $\ff$ to decrease the KL divergence between the distribution of particles and the target $p(x)$ as fast as possible. 
This can be solved by the following basic observation shown in \citet{liu2016stein}: 
assume $x\sim q$ and $q_{[\epsilon\ff]}$ is the distribution of $x' = x+\epsilon \ff(x)$, then 
$$
 \KL(q_{[\epsilon \ff]} \parallel p) =  \KL(q \parallel p) - \epsilon~ \E_{x\sim q} [\stein^\top_x \ff(x)] + O(\epsilon^2), 
$$
where $\steinpx$ is a linear operator, called Stein operator, that acts on function $\ff$ via 
\begin{align}
\stein^\top_x \ff(x) ~~\eqdef ~~  \nabla_x \log p(x)^\top \ff(x) + \nabla_x^\top \ff(x). 
\label{def:stein_op}
\end{align}
This result suggests that the decreasing rate of KL divergence when applying transform $x'=x+\epsilon\ff(x)$ 
is dominated by $\E_{x\sim q} [\steinp^\top_x \ff(x)]$ when the step size $\epsilon$ is small. 
In a special case when $q=p$, Stein operator draws connection to Stein's identity, which shows 
$$
\E_{x\sim q} [\stein^\top_x \ff(x)] =0~~~~~ \text{when~~ $p = q$.}
$$
This is expected because as $q=p$, the decreasing rate of KL divergence must be zero
for all velocity fields $\ff$. 

Given a candidate set $\F$ of velocity fields $\ff$, we should choose the best $\ff$ to maximize the decreasing rate, 
\begin{align}
\label{equ:ksd_prob}
\S(\qo ~||~ p) ~~\eqdef ~~ \max_{\ff \in \F} \bigg\{ \E_{x\sim \qo} [\stein^\top_x \ff(x)] \bigg\}, 
\end{align}
where the maximum decreasing rate $\S(\qo~||~p)$ is called the \emph{Stein discrepancy} 
between $q$ and $p$. 
Assume $\F$ includes $\pm\ff$ for $\forall \ff\in\F$, then \eqref{equ:ksd_prob} is equivalent to maximizing the absolute value of $\E_q[\stein^\top_x \ff]$ and $\S(\qo ~||~ p)$ must be non-negative for any $q$ and $p$.  
In addition, If $\F$ is taken to be rich enough, $\S(\qo ~||~p) = 0$ only if there exists no velocity field $\ff$ that can decrease the KL divergence between $p$ and $\qo$, which must imply $p=\qo$. 

To make $\F$ computationally tractable,
\citet{liu2016stein} further assumed that $\F$ is the unit ball of a vector-valued RKHS $\H = \H_0\times \cdots \times \H_0$, where each $\H_0$ is a scalar-valued RKHS associated with a positive definite kernel $k(x,x')$.
In this case, \citet{liu2016kernelized} showed that the optimal solution of (\ref{equ:ksd_prob}) is $\ff^*/|| \ff^*||$, where 
\begin{align}
\ff^*(\cdot) & = \E_{x\sim \qo} [\steinpx k(x,\cdot)] \nonumber \\
              & = \E_{x\sim \qo}[ \nabla_x \log p(x) k(x,\cdot) + \nabla_x k(x,\cdot)].
\label{equ:phistar}
\end{align}
This gives the optimal update direction within RKHS $\H$. 
By starting with a set of initial particles and then repeatedly applying this update with $q$ replaced by the empirical distributions of the particles, we obtain the SVGD algorithm: 
\begin{align}
\label{equ:svgd_update}
& x^\ell \gets x^\ell + \epsilon \ff^*(x^\ell), ~~~ ~~\text{} ~~~~ \forall \ell = 1,\ldots, n, \\ 
& \ff^*(\cdot) =\frac{1}{n}\sum_{\ell=1}^n[\nabla_{x^{\ell}} \log p(x^{\ell}) k( x^{\ell},~\cdot) + \nabla_{x^{\ell}} k(x^{\ell},~\cdot)]. \nonumber
\end{align}
The two terms in $\ff^* (x)$ play different roles: 
the first term with the gradient $\nabla_x \log p(x)$ drives the particles toward the high probability regions of $p(x)$,
while the second term with $\nabla_{x^\ell} k(x,x^\ell)$ serves as a repulsive force to encourage diversity as shown in \citet{liu2016stein}. 
It is easy to see from \eqref{equ:svgd_update} that SVGD reduces to standard gradient ascent for maximizing $\log p(x)$ 
(i.e., maximum a posteriori (MAP)) when there is only a single particle $(n=1)$. 

\paragraph{Discriminative Stein Discrepancy} 
One key requirement when selecting the function space $\H$ is that it should be rich enough to make Stein discrepancy discriminative, 
that is, 
\begin{align}\label{iff}
\S(q ~ || ~ p) = 0\text{~~~~~~ implies ~~~~~~ }q = p. 
\end{align}
This problem has been studied in various recent works, 
including \citet{liu2016kernelized, chwialkowski2016kernel, oates2016convergence, gorham2017measuring}, all of which require  
$\H$ to be an universal approximator in certain sense. 
In particular, the condition required in \citet{liu2016kernelized}
is that the kernel $k(x,x')$ should be \emph{strictly integrally positive definite} in the sense that 
\begin{equation}\label{ipd}
\int g(x) k(x,x') g(x') dx dx' > 0, 
\end{equation}
for any nonzero function  $g$ with  $0 < || g||_2^2 < \infty$. 
Many widely used kernels, including Gaussian RBF kernels of form $k(x,x') = \exp(-\frac{1}{h}{|| x - x'||^2_2})$, are strictly integrally positive definite.

%% file: tex/theory.tex
\section{SVGD for Graphical Models }

The goal of our work is to
extend SVGD to approximate 
high dimensional probabilistic graphical models of form 
\begin{align}\label{equ:graphical}
p(x) \propto 
\exp\big [ \sum_{s \in \mathcal{S}} \psi(x_s) \big ], 
\end{align}
where $\mathcal{S}$ is a family of index sets $s\subseteq \{1,\ldots, d\}$, 
and $x_{s} = [x_i]_{i \in s}$ represents the sub-vector of $x$ over index set $s$. 
Here the clique        set $\mathcal{S}$ specifies the \emph{Markov structure} of the graphical model: 
for each variable (or node) $i$, 
its Markov blanket (or neighborhood) $\MB_i$ is the set of nodes that co-appears in at least one clique $s \in \mathcal{S}$, that is,  
$
\MB_i := \cup \{ s  \colon s \in \mathcal{S}, ~ s \ni i \}\setminus \{i\}.
$
Related, we define the \emph{closed} neighborhood (or clique) of node $i$ to be 
$\C_i := \MB_i \cup \{i\}$. 
The local Markov property guarantees that a variable $x_i$ is 
conditionally independent of all other nodes given its Markov blanket ${\MB_i}$. 

The Markov structure is also reflected in gradient $\nabla \log p$: 
$$
\partial_{x_i} \log p(x) 
= \sum_{ s \ni i } \partial_{x_i} \psi(x_s),
$$
where $\partial_{x_i}$ denotes the partial derivative with respect to $x_i$, the $i$-{th} component of $x$. 
This suggests that the gradient evaluation of node $i$ only requires information from its closed neighborhood set $\C_i$; this property makes  gradient-based distributed inference methods a possibility. 

Unfortunately, when directly applying SVGD to graphical models,  
the Markov structure is not inherited in the SVGD update 
(\ref{equ:svgd_update}), because typical kernel functions do not have the same (additive) factorization structure as $\partial_{x_i} \log p(x)$. 
Take the commonly used Gaussian RBF kernel $k(x,x') = \exp(-\frac{1}{h}||x - x' ||_2^2)$ as an example. 
Because $k(x,x')$ involves all the coordinates of $x$,
the $i$-th coordinate of 
the Stein variational gradient $\ff^*(x)$ in \eqref{equ:svgd_update} depends on all the other nodes, 
including those out of the Markov blanket of node $i$. 
This makes it infeasible to apply vanilla SVGD in distributed settings because the update of node $i$ requires information from all the other nodes. 

\begin{figure}[t]
\centering
\begin{tabular}{cc}
\raisebox{1.0em}{\rotatebox{90}{\small Estimated Variance}}
\includegraphics[width=0.24\textwidth]{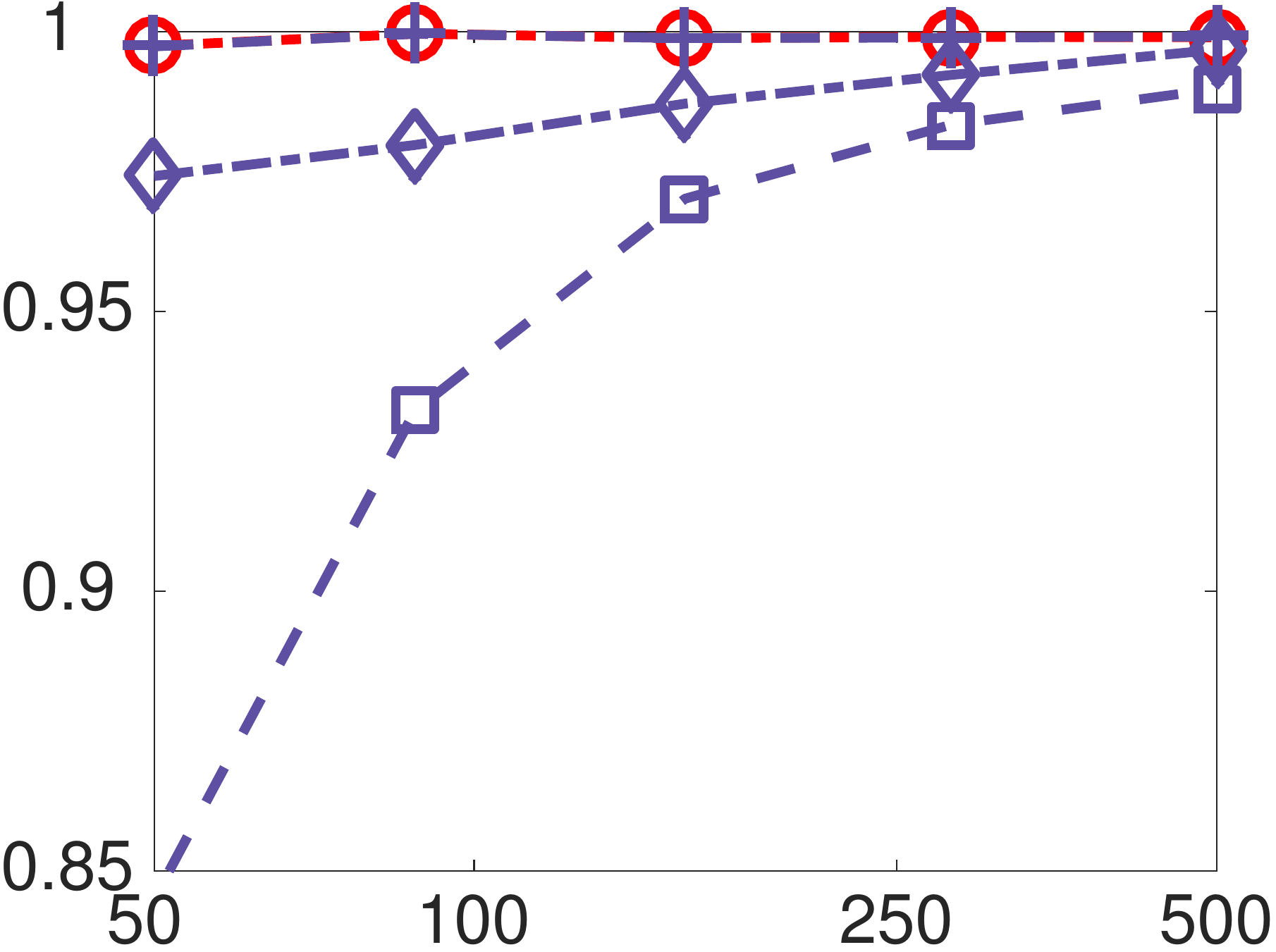} &
\raisebox{2.0em}{\includegraphics[width=0.15\textwidth]{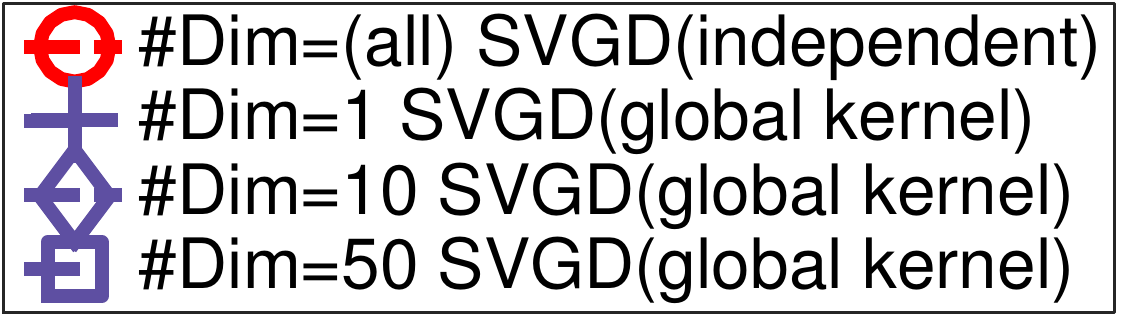}} \\
~~~~~Number of Particles\\
\end{tabular}
\caption{Estimating the variance using SVGD when $p(x)$ is the standard multivariate Gaussian distribution $\normal(0, I_d)$ of different dimensions $d$ (the true variance is $\sigma = 1$).  
}
\label{fig:iso-gaussian}
\end{figure}

More importantly, this global dependency introduced by kernel functions may also lead to poor performance 
in high dimensional models. 
To illustrate this, 
we consider a simple example when the distribution $p(x)$ of interest is fully factorized, 
that is, $p(x) =  \prod_{i=1}^d p_i(x_i)$, 
where each node $x_i$ is independent of all the other nodes. 
In this case, we find that SVGD with the standard Gaussian RBF kernel (see {\em SVGD (global kernel)} in Figure~\ref{fig:iso-gaussian})  
requires increasingly more particles in order to estimate the variance accurately as the dimension $d$  increases;  
this is caused by the additional (and unnecessary) dependencies introduced by the use of the global RBF kernel. 

In this particular case, an obviously better approach 
is to apply SVGD with RBF kernel on each of the one-dimensional marginal $p_i(x_i)$ individually (see {\em SVGD (independent)} in Figure~\ref{fig:iso-gaussian})); 
 this naturally leverages the fully factorized structure of $p(x)$, and makes the algorithm immune to the curse of dimensionality because the RBF kernel is applied on an individual variable each time. 
 Therefore, exploiting the sparse dependency structure 
 can significantly improve the performance in high dimensions. 
The key question, however, is how we can extend the idea beyond the fully factorized case. 
This motivates our graphical SVGD algorithm as we show in the sequel. 

\subsection{Graphical SVGD} 
In order to extend the above example, 
we observe that running 
SVGD on each marginal distribution $p_i(x_i)$ independently 
can be viewed as 
a special SVGD applied on the joint distribution $p(x)=\prod_i p_i(x_i)$, but updating each coordinate $x_i$ using its own kernel $k_i(x,x^\prime)$ that only depends on the $i$-th coordinate, that is,  $k_i(x, x^\prime) := k_i(x_i, x_i^\prime).$ 
An intuitive way to extend this to $p(x)$ with more general Markov structures is to 
update each $x_i$ with a local kernel function $k_i(x,x') := k_i(x_{\C_i}, x_{\C_i}')$ that depends only on the closed neighborhood $\C_i$ of node $i$, that is, 
\begin{align}
& x_i^\ell \leftarrow x_i^\ell + \epsilon \phi^*_i(x^\ell), ~~~~~~ 
\forall i \in [{d}], ~ \ell \in [{n}]  \label{equ:gsvgd}\\
& \phi^*_i(x) :=  \frac{1}{n} \sum_{\ell=1}^n [ s_i(x^\ell) k_i( x_{\C_i}^\ell, x_{\C_i}) 
 + \partial_{x_i^\ell} k_i(x_{\C_i}^\ell, x_{\C_i})], \notag
\end{align}
where $s_i(x) = \partial_{x_i} \log p(x)$. This procedure, which we call graphical SVGD  (see also Algorithm~\ref{alg:graphical_svgd}), provides a simple way to efficiently integrate the graphical structure of $p(x)$ into SVGD. As we demonstrate in our experiments,  it significantly improves the performance in  high dimensional, sparse graphical models. 
In addition, it yields a communication-efficient distributed message passing form, 
since the update of $x_i$ only requires to access 
the neighborhood variables in ${\C_i}$.   

Is this graphical SVGD update theoretically sound? 
The key difference between \eqref{equ:gsvgd} and the original SVGD  \eqref{equ:svgd_update} includes: 
i) graphical SVGD uses 
a different kernel $k_i(x,x')$ for each coordinate $x_i$;  and 
ii) each kernel $k_i(x,x') = k_i(x_{\C_i}, x_{\C_i}')$ only depends on the closed neighborhood $\C_i=\{i\}\cup \N_i$ of node $i$. 
We justify these two choices theoretically in Section~\ref{sec:manykernels} and Section~\ref{sec:structuredkernels}, respectively. 

\subsection{Stein Discrepancy with Coordinate-wise Kernels}
\label{sec:manykernels}
 Using coordinate-wise kernel $k_i(x,x')$ can be simply viewed as 
  taking the space  $\H$ in  the functional optimization \eqref{equ:ksd_prob}  
to be a more general product space 
$\H = \H_1 \times \cdots \times \H_d,$
where each individual RKHS $\H_i$ is related to a different kernel $k_i(x,x')$; 
the original SVGD can be viewed as a special case of this when all the kernels $k_i(x,x'),$ $\forall i\in [d]$ equal.  
The following result extends Lemma~3.2 and Theorem~3.3 of \citet{liu2016stein} 
to take into account coordinate-wise kernels. 
\begin{thm}\label{thm:optimal}
Let $\H_i$ be the (scalar-valued) RKHS related to kernel $k_i(x,x')$, and $\H = \H_1 \times \cdots \times \H_d$ their product RKHS consisting of $\ff = [\phi_1, \ldots, \phi_d]^\top$ with norm $||\ff||_{\H}^2=\sum_i ||\phi_i||_{\H_i}^2$,  $\forall \phi_i \in \H_i$.  
Taking $\F = \{\ff  \in \H \colon  ||\ff ||_\H \leq 1\}$ in the optimization problem in \eqref{equ:ksd_prob},  then the optimal solution is
\begin{align}
& \ff^*(\cdot)= \E_{x \sim q}[\steinpx \circ \vv k(x,\cdot)], \label{equ:phiddd} \\
& \text{with} ~ \steinpx \circ \vv k(x, \cdot) \overset{def}{=} [ \steinp_{x_1} k_1(x, \cdot) , \cdots, \steinp_{x_d} k_d(x, \cdot)  ]^\top \notag, 
\end{align}
where $\circ$ denotes the entrywise product between $\steinpx$ and $\vv k = [k_1, \ldots, k_d]$. 
Further, the related Stein discrepancy in \eqref{equ:ksd_prob} equals $\S(q ~||~ p) = ||\ff^* ||_\H$. 
\end{thm}

This result suggests 
that updates of form $x_i' \gets x_i' + \epsilon \E_{x\sim q}[\steinp_{x_i} k_i(x,x')]$  
yields the fastest descent direction of KL divergence within $\H$, justifying 
the graphical SVGD update in \eqref{equ:gsvgd}. 

The following result studies the properties of the Stein discrepancy related to coordinate-wise kernels $\vv k = [k_1, \ldots, k_d]$, showing that $\S(q~||~p)$ is discriminative if all the kernels $k_i(x,x')$ are strictly integrally positive definite, 
a result that generalizes Proposition~3.3 of \citet{liu2016kernelized}. 

\begin{algorithm}[t] 
\caption{Graphical Stein Variational Gradient Descent} 
\label{alg:graphical_svgd}
\begin{algorithmic} 
\STATE {\bf Input:} A graphical model  $p(x)$ with Markov blanket $\MB_i $ for node $i$; $\C_i=\MB_i \cup \{i\};$
a set of local kernels $k_i(x_{\C_i}, x_{\C_i}')$, and initial particles $\{x^{\ell,0}\}_{\ell=1}^n$; 
step size $\epsilon$. 
\STATE {\bf Goal:} A set of particles $\{x^\ell\}_{\ell=1}^n$ that approximates $p(x)$
\FOR{iteration $t$} 
\FOR{node $i$} \vspace{1em}
\STATE
$
 \begin{aligned}
& x_i^{\ell, t+1} \leftarrow x_i^{\ell,t} + \epsilon \phi_i^*(x^{\ell,t})  \\
& \text{where $\phi^*_i(x)$ is defined in \eqref{equ:gsvgd} (with $x^\ell = x^{\ell,t}$).}
\end{aligned}
$
\ENDFOR
\ENDFOR
\end{algorithmic} 
\end{algorithm}
\vspace{-10pt}

\begin{thm}\label{lem:graphical_svgd_update}
Following Theorem~\ref{thm:optimal}, the Stein discrepancy $\S(q~||~p)$ 
with kernels $\vv k = [k_1, \ldots, k_d]$ satisfies  
$$
\S(q~||~p)^2 = \sum_{i=1}^d \E_{x,x'\sim q}[\steinp_{x_i} \steinp_{x_i'} k_i(x,x')]. 
$$
Further, Assume both $ p(x)$ and $ q(x)$ are positive and differentiable densities.
Denote by $\mathcal Q_\vx$ the Stein operator of distribution $q$. 
If Stein's identity $\E_{x\sim q}[\mathcal Q_{x_i} k_i(x,x')] = 0$, $\forall x'\in \X$ holds for all the kernels $k_i$, $\forall i \in [d]$, we have 
\begin{align}\label{equ:deltaq}
\S(q~||~p)^2 = \sum_{i=1}^d \E_{x,x'\sim q} [ \delta_i(x) k_i(x,x') \delta_i(x')], 
\end{align}
where $\delta_i(x) = \nabla_{x_i} \log p(x_i|x_{\neg i}) - \nabla_{x_i} \log q(x_i | x_{\neg i})$,
with $\neg i = \{1,\ldots, n\}\setminus \{i\}$. 

Assume $||q \delta_i||_2^2 < \infty$, $\forall i \in [d]$. 
If all the kernels $k_i(x, x')$ are strictly integrally positive definite in the sense of \eqref{ipd}, then
$\S(q~||~p)=0$ implies $q=p$. 
\end{thm}

%% file: tex/theory_graphical.tex
\subsection{Stein Discrepancy with Local Kernels}
\label{sec:structuredkernels}  
Theorem~\ref{lem:graphical_svgd_update} requires every kernel $k_i(x,x')$  to be strictly integral positive definite 
to make Stein discrepancy discriminative. 
However, in our graphical SVGD, each kernel function $k_i(x,x') := k_i(x_{\C_i}, x_{\C_i}')$
only depends on a subset of the variables and can be easily seen to be not strictly integrally positive definite in the sense of \eqref{ipd}. 
As a result, the related Stein discrepancy is no longer discriminative in general. 

Fortunately, as we show in the following, 
once each $k_i(x_{\C_i}, x_{\C_i}')$ is strictly integrally positive definite w.r.t. its own local variable domain $x_{\C_i}$, that is, 
\begin{align}\label{equ:cipsd}
\int g(x_{\C_i}) k_i(x_{\C_i}, x_{\C_i}') g(x_{\C_{i}}') dx_{\C_i} dx_{\C_i}' > 0
\end{align}
for any function $g(x_{\C_i})$ with  $0< ||g||_2^2 < \infty$, then 
a zero Stein discrepancy $\S(q~||~p) = 0$  guarantees to match the conditional distributions: 
$$q( x_i ~|~ x_{\N_i}) = p(x_i ~|~ x_{\N_i})
~~~~~\text{for any $i \in [d]$}.
$$ 
Although this in general does not necessarily imply that the joint distribution equals ($q = p$) (unless $q$ is guaranteed a priori to have the same Markov structure as $p$), 
it suggests that graphical SVGD
captures important perspectives of the target distribution,  
especially in terms of the quantities related to the local dependency structures.  





\begin{thm}\label{thm:graphical_svgd_update}
Assuming that $p(x)$ is a graphical model in which the  Markov neighborhood of node $i$ is  $\N_i$  and $\C_i = \MB_i \cup \{i\}$,
and that $k_i(x,x') = k_i(x_{\C_i}, x_{\C_i}')$, ~$\forall i \in [d]$, then 
 \eqref{equ:deltaq} reduces to  
\begin{align*} 
\S(q~||~p)^2 = \sum_{i=1}^d 
\E_{x,x'\sim q} [ \delta_i(x_{\C_i}) k_i(x_{\C_i},x_{\C_i}') \delta_i(x_{\C_i}')], 
\end{align*}
where $\delta_i(x_{\C_i}) = \nabla_{x_i} \log p(x_i | x_{\N_i}) - \nabla_{x_i} \log q(x_i | x_{\N_i})$.
Further, assume $g(x_{\C_i}) := q(x_{\C_i}) \delta_i(x_{\C_i})$ and $||g||_2^2 <\infty$. 
If each kernel $k_i(x_{\C_i},x_{\C_i})$ is integrally strictly positive definite w.r.t. variable $x_{\C_i}$ as defined in \eqref{equ:cipsd}, we have 
$$
\S(q~||~p) = 0 ~~~~\text{iff}~~~~ q(x_i | x_{\N_i})  = p(x_i | x_{\N_i}), ~~\forall i \in [d]. 
$$
\end{thm}

%
%

Generally speaking, matching the condition distributions shown in Theorem~\ref{thm:graphical_svgd_update} 
does not guarantee to match the joint distributions ($p=q$). 
A simple counter example is when $p(x)$ is fully factorized, $p(x)= \prod_i p_i(x_i)$ and $\N_i = \emptyset$, in which case 
matching the singleton marginals $p(x_i) = q(x_i)$, $i\in [d]$ does not imply $p(x)=q(x)$ jointly since one can construct infinite many distributions with the same singleton marginals using the Copula method \citep{nelsen2007introduction}. 

Therefore, graphical SVGD can be viewed as a \emph{partial} reconstruction of the target distribution $p$,
which retains the local dependency structures while ignoring the long-range relations which are inherently difficult to infer due to the curse of high dimensionality. 
Focusing on the local dependency structures makes the problem easier and tractable, and also of practical interest since local marginals are often what we care in practice. 
Therefore, graphical SVGD can be viewed as an interesting hybrid of deterministic approximation (via the use of local kernels) and particle approximation (by approximating $q$ with the empirical distributions of the particles). 
%

%% file: tex/experiments.tex
\begin{figure*}[t]
\centering
\begin{tabular}{ccccc}
\raisebox{1.4em}{\rotatebox{90}{\scriptsize Log10~MSE}}
\includegraphics[width=0.18\textwidth]{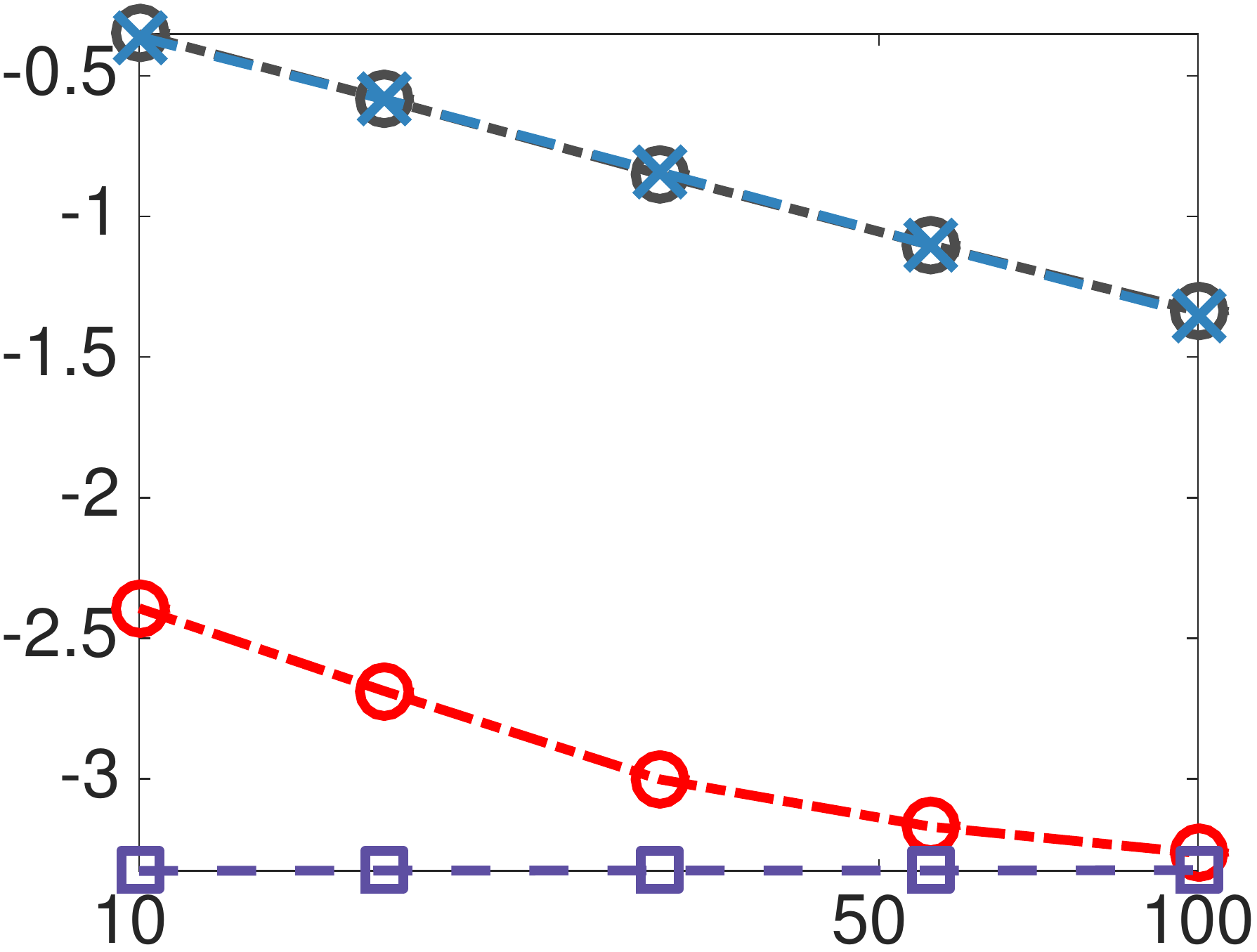} &
\hspace{-1em}
\raisebox{1.4em}{\rotatebox{90}{\scriptsize Log10~MSE}}
\includegraphics[width=0.18\textwidth]{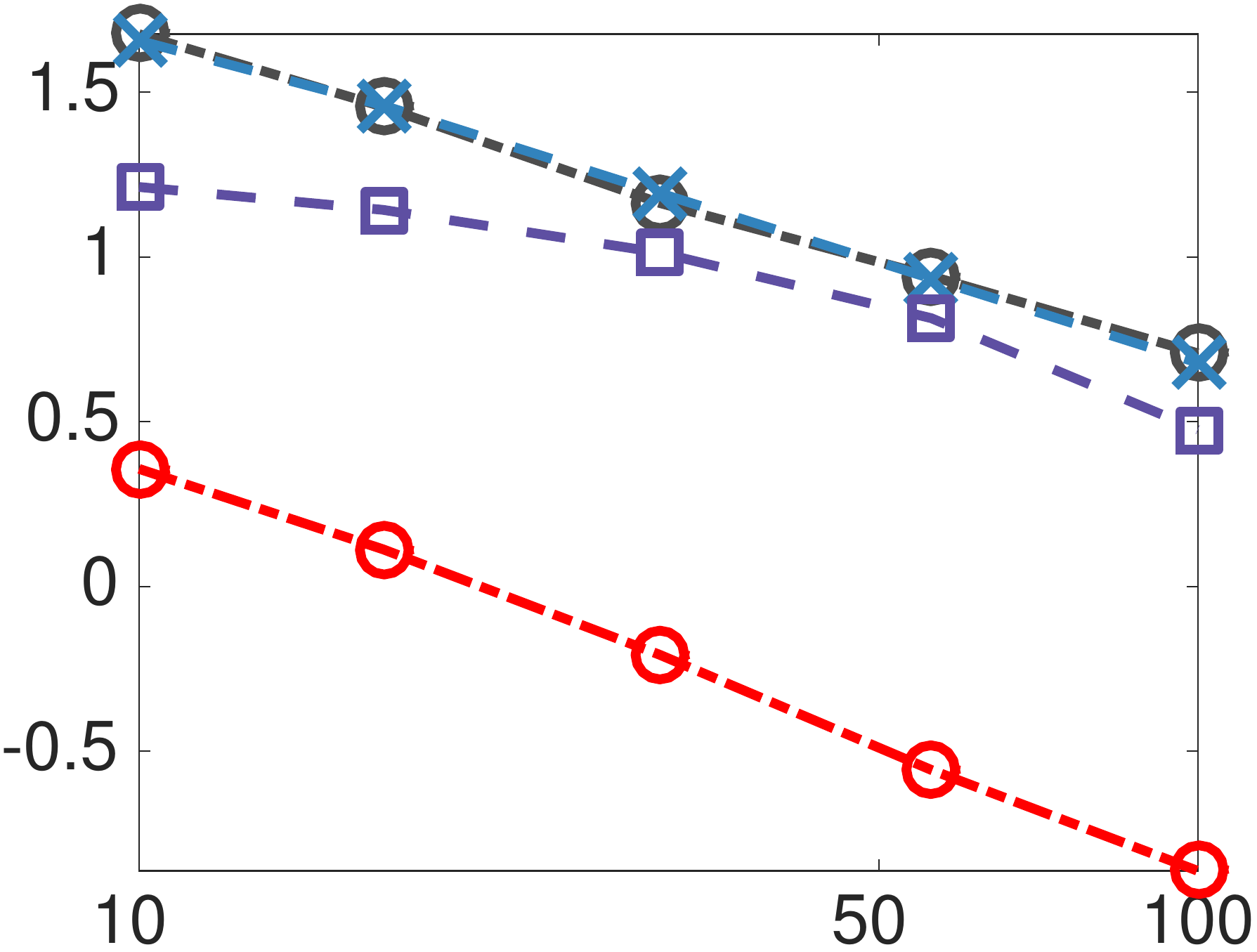} &
\hspace{-1em}
\raisebox{1.4em}{\rotatebox{90}{\scriptsize Log10~MMD}}
\includegraphics[width=0.18\textwidth]{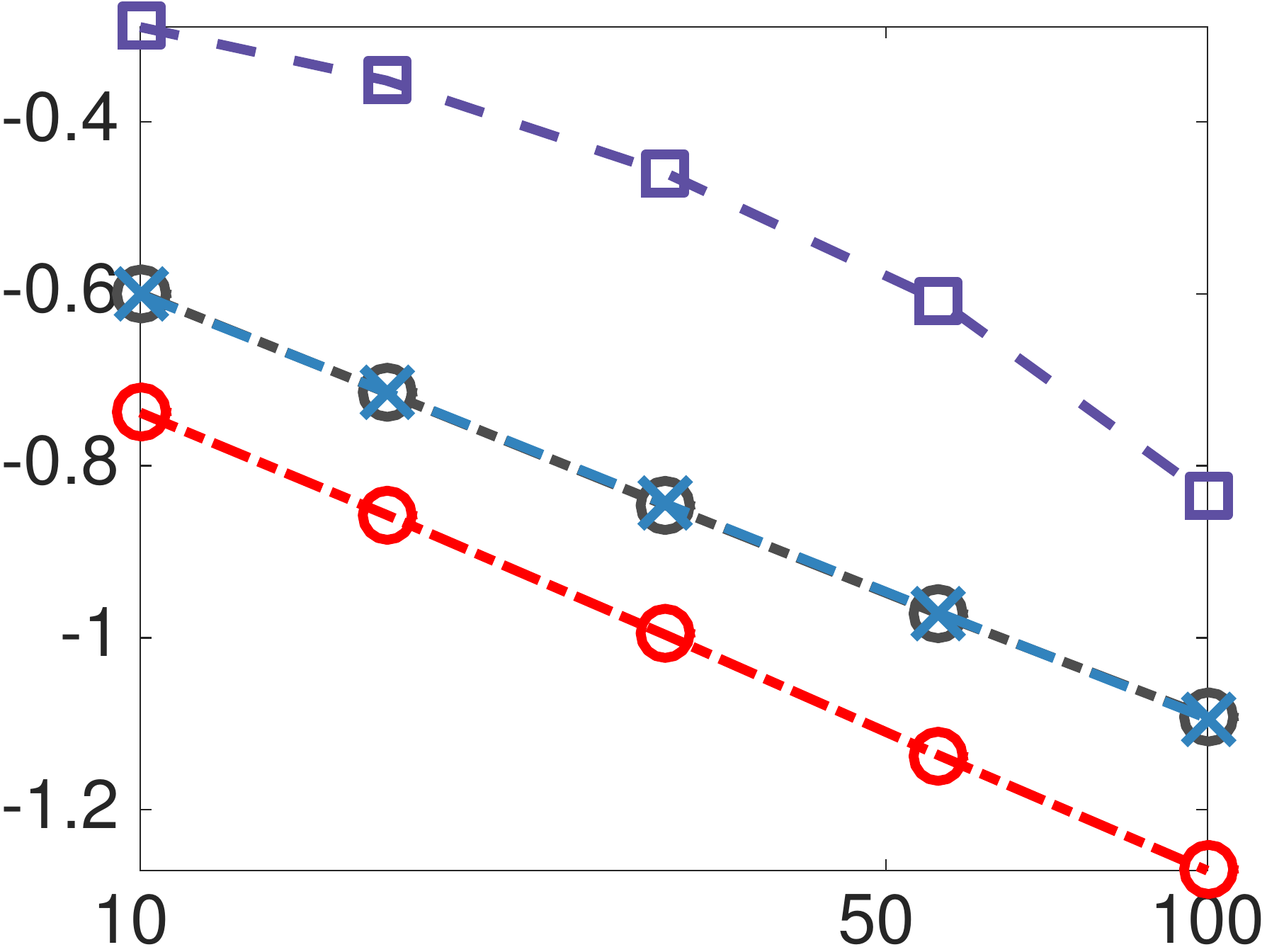} &
\hspace{-1.6em}
\raisebox{4em}{\includegraphics[width=0.12\textwidth]{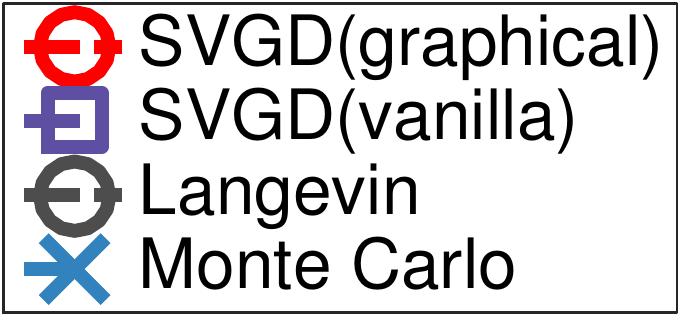}} & 
\raisebox{1.4em}{\rotatebox{90}{\scriptsize Log10~MMD}}
\includegraphics[width=0.18\textwidth]{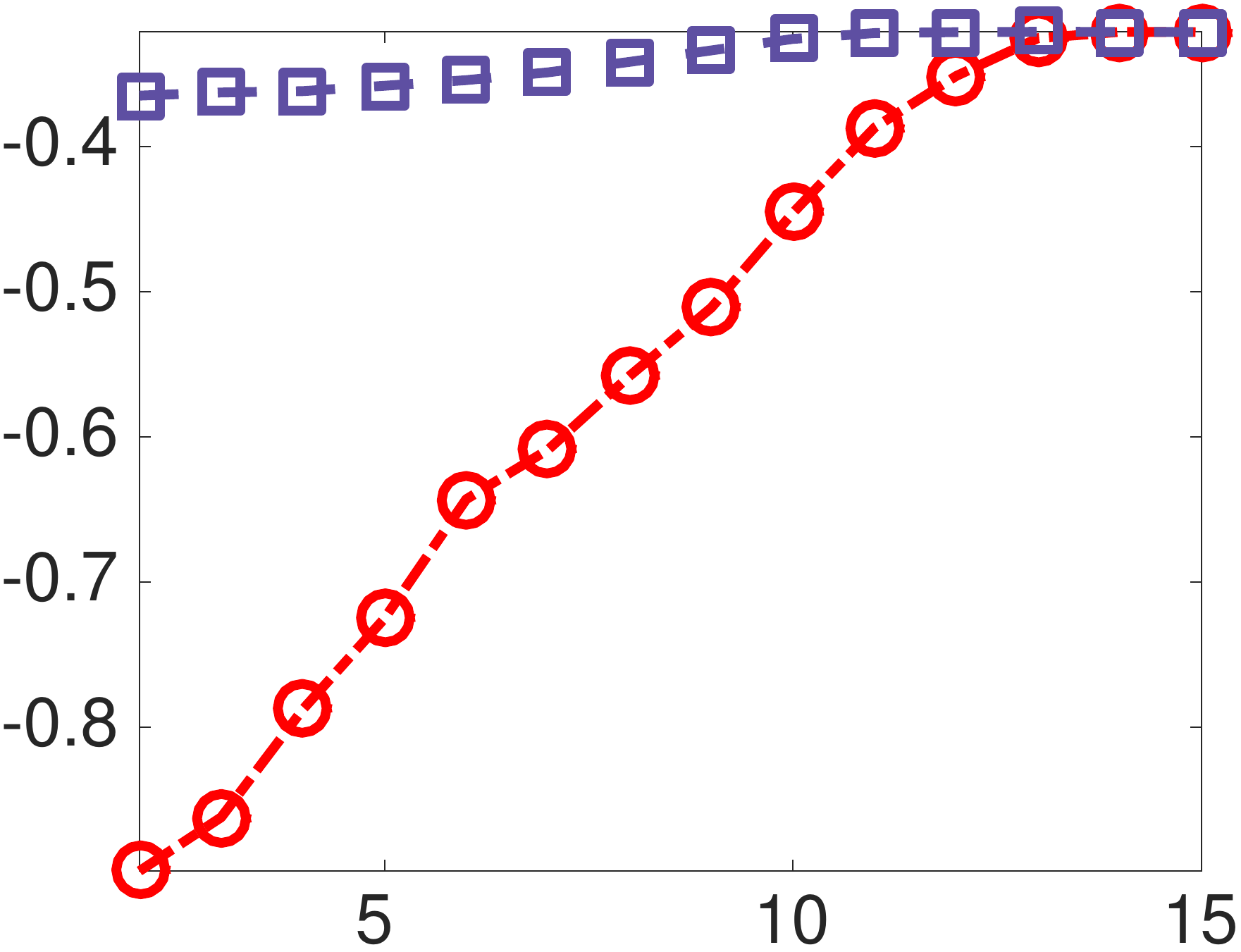} \\
\scriptsize Number of samples &\scriptsize  Number of samples  &\scriptsize Number of samples & &   \scriptsize Connectivity Radius ($r$)  \\
(a) 
\small Estimating $\E[x_i]$ & (b) \small Estimating $\E[x_i^2]$ & \small (c) MMD vs. $n$ & & (d) \small MMD vs. graph sparsity \\
\end{tabular} 
\caption{
(a)-(b): Results of Gaussian MRFs on a $10\times 10$ 4-neighborhood 2D grid, 
evaluated using the MSE for estimating the mean (a), the MSE for estimating the second order moments (b), 
and the maximum mean discrepancy between the particle and the true distribution (c). 
In (d), we show the performance of graphical SVGD and vanilla SVGD when the connectivity of the graph increases (with a fixed sample size of n=20). 
All results are averaged over 50 random trials.}
\label{fig:gaussian_mrfs}
\end{figure*}

\begin{figure*}[t]
\centering
\begin{tabular}{cccc}
\raisebox{1.4em}{\rotatebox{90}{\scriptsize Log10~MSE}}
\includegraphics[width=0.18\textwidth]{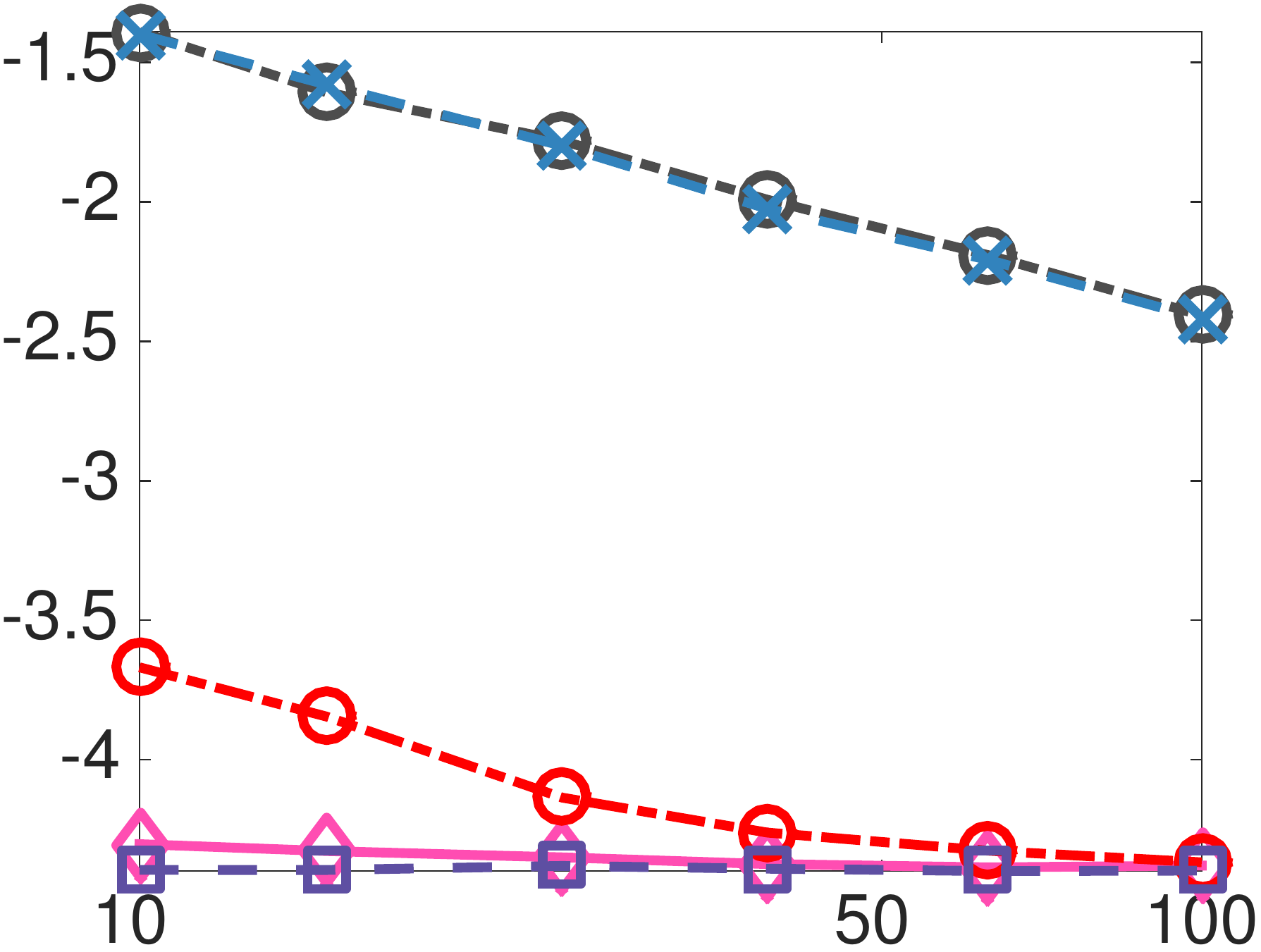}~~ &
\raisebox{1.4em}{\rotatebox{90}{\scriptsize Log10~MSE}}
\includegraphics[width=0.18\textwidth]{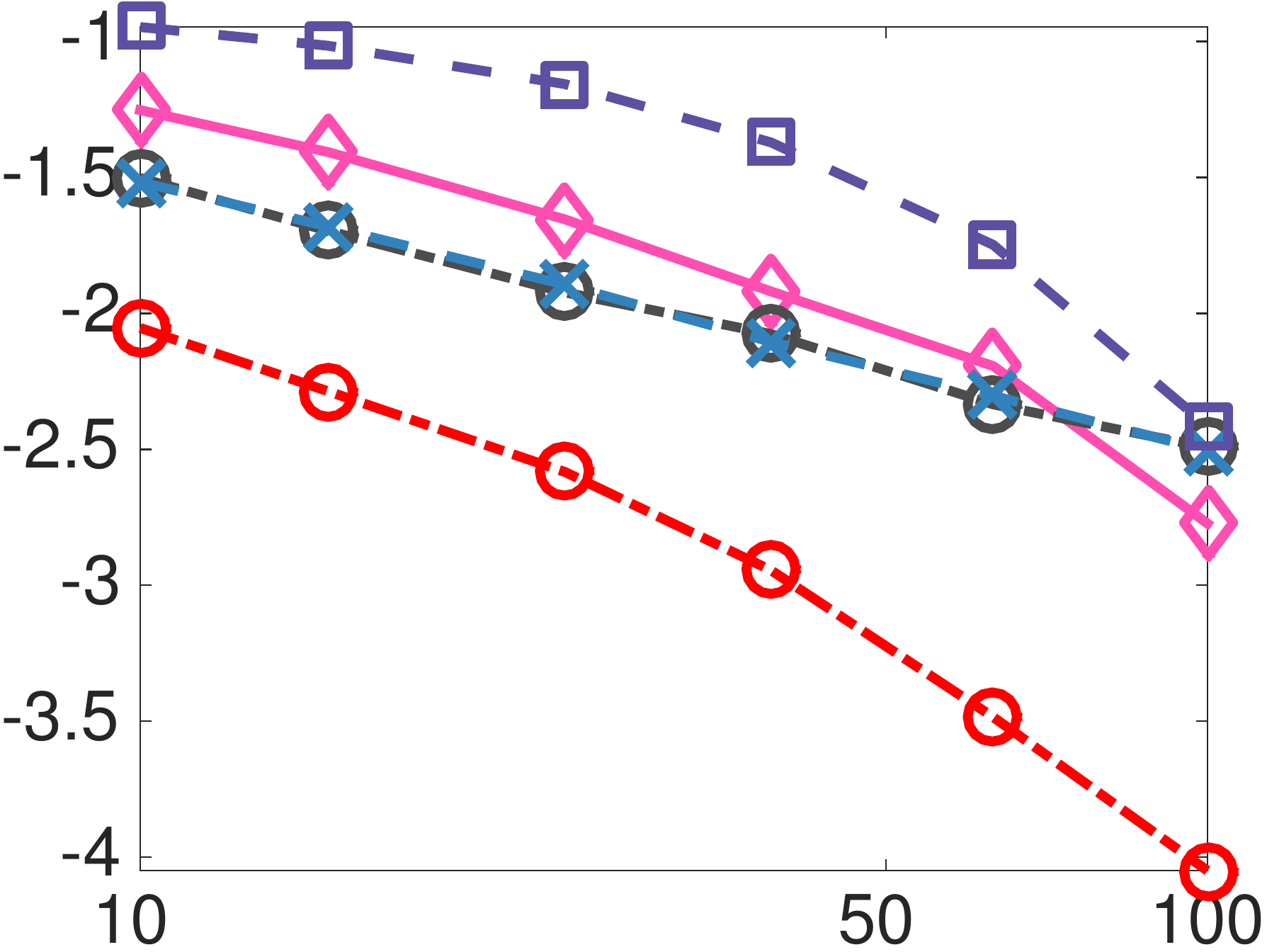} ~~&
\raisebox{1.4em}{\rotatebox{90}{\scriptsize Log10~MMD}}
\includegraphics[width=0.18\textwidth]{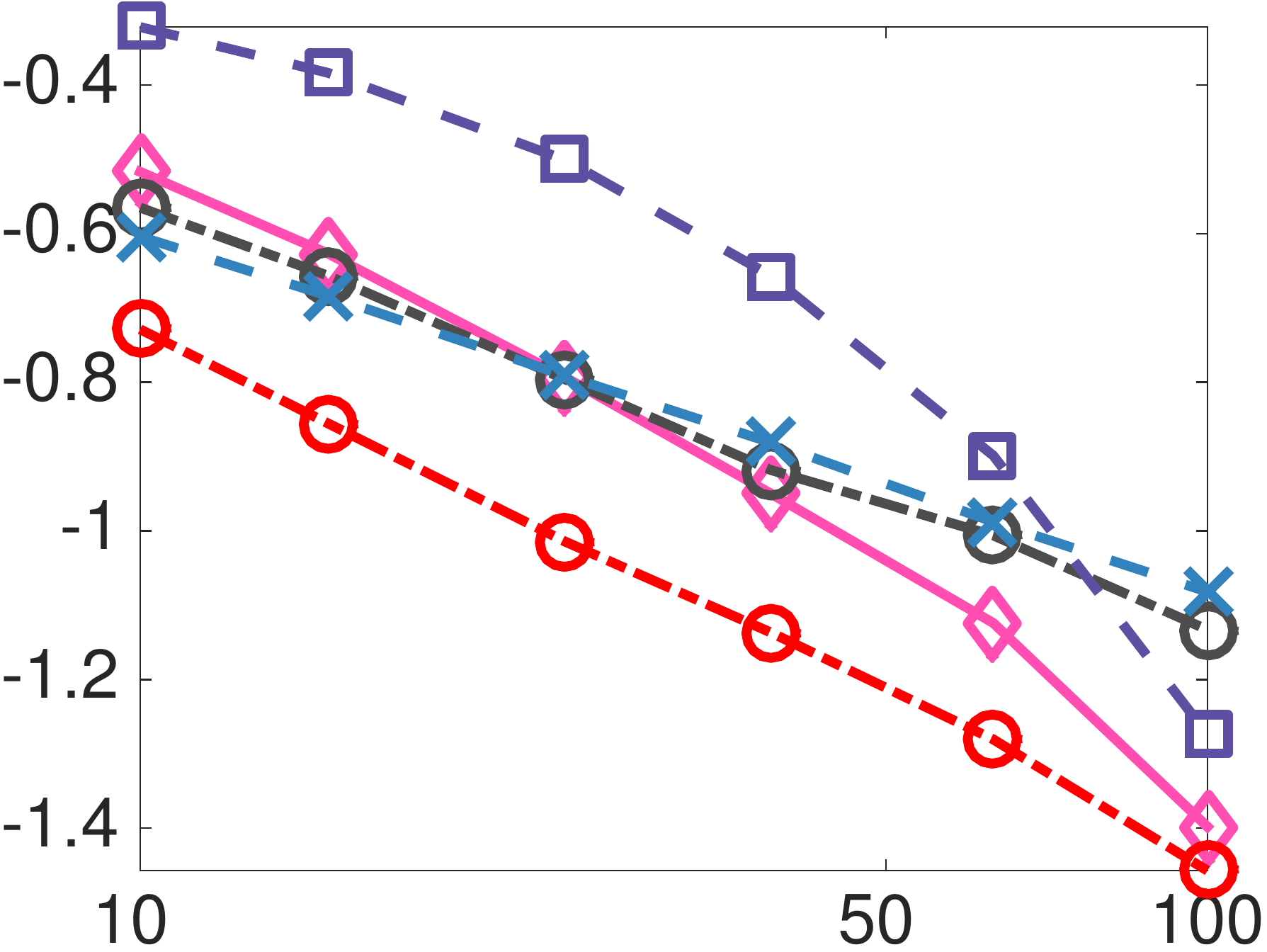} ~~&
\raisebox{1em}{\includegraphics[width=0.14\textwidth]{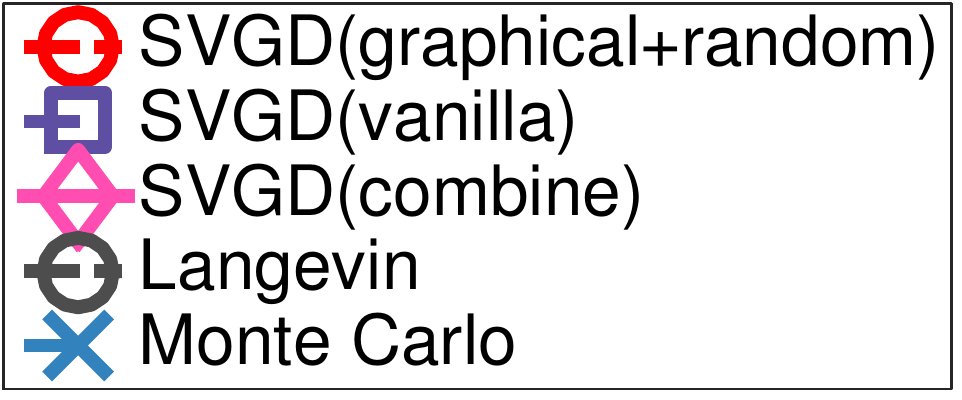}} ~~\\
\scriptsize Number of samples & \scriptsize Number of samples  & \scriptsize Number of samples & \\
(a) \small Estimating $\E[x_i]$ & (b) \small Estimating $\E[x_i^2]$ & \small (c) MMD vs. $n$ &  \\
\end{tabular}
\caption{Results on 50 dimensional fully connected Gaussian MRFs. 
In this case, \textit{SVGD (graphical+random)} uses $k_i(x,x') = k_i(x_{\mathcal D_i}, x_{\mathcal D_i}')$ 
where
$\mathcal D_i$ consists of node $i$ and four neighbors of $i$ randomly selected at each iteration. 
\textit{SVGD (combine)} uses $k_i(x,x') =  (k_i(x_{\mathcal D_i}, x_{\mathcal D_i}') +  k_0(x, x'))/2$ where $k_0(x,x')$ is the regular global RBF kernel.
}
\label{fig:gaussian_dense}
\end{figure*}

\begin{figure*}[t]
\centering
\begin{tabular}{ccc}
\includegraphics[height=0.18\textwidth]{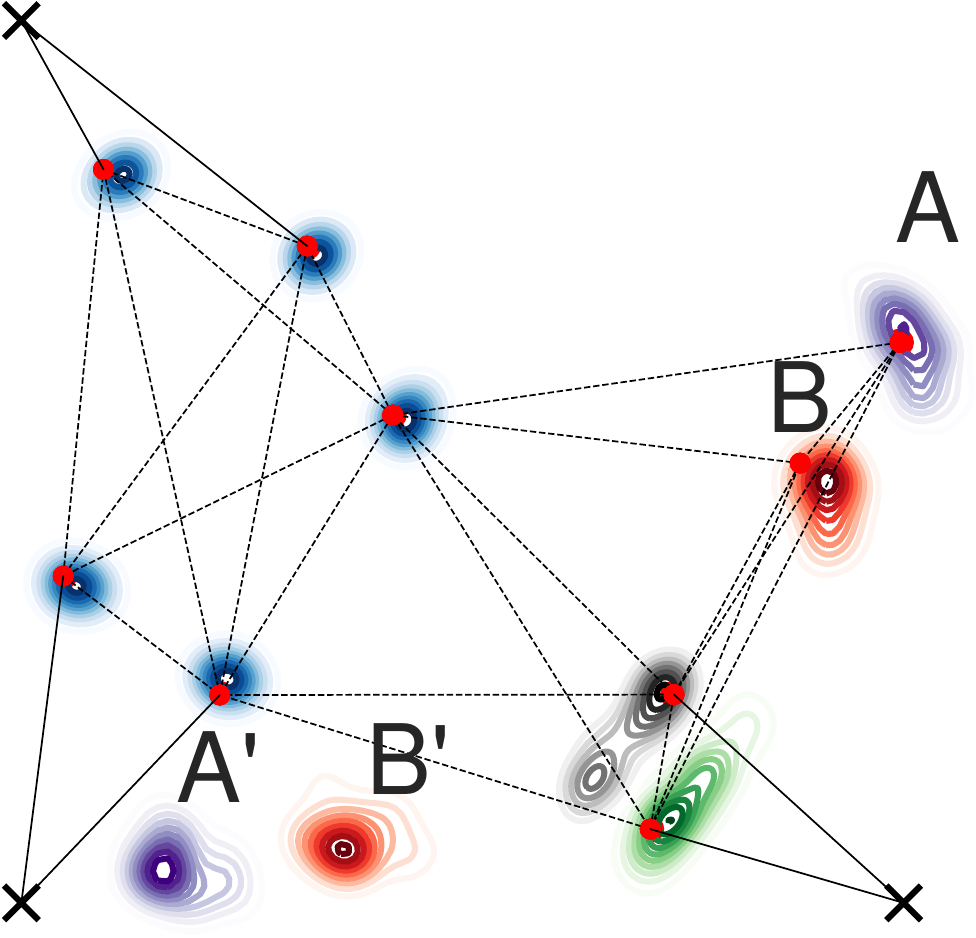}  &
~~\includegraphics[height=0.18\textwidth]{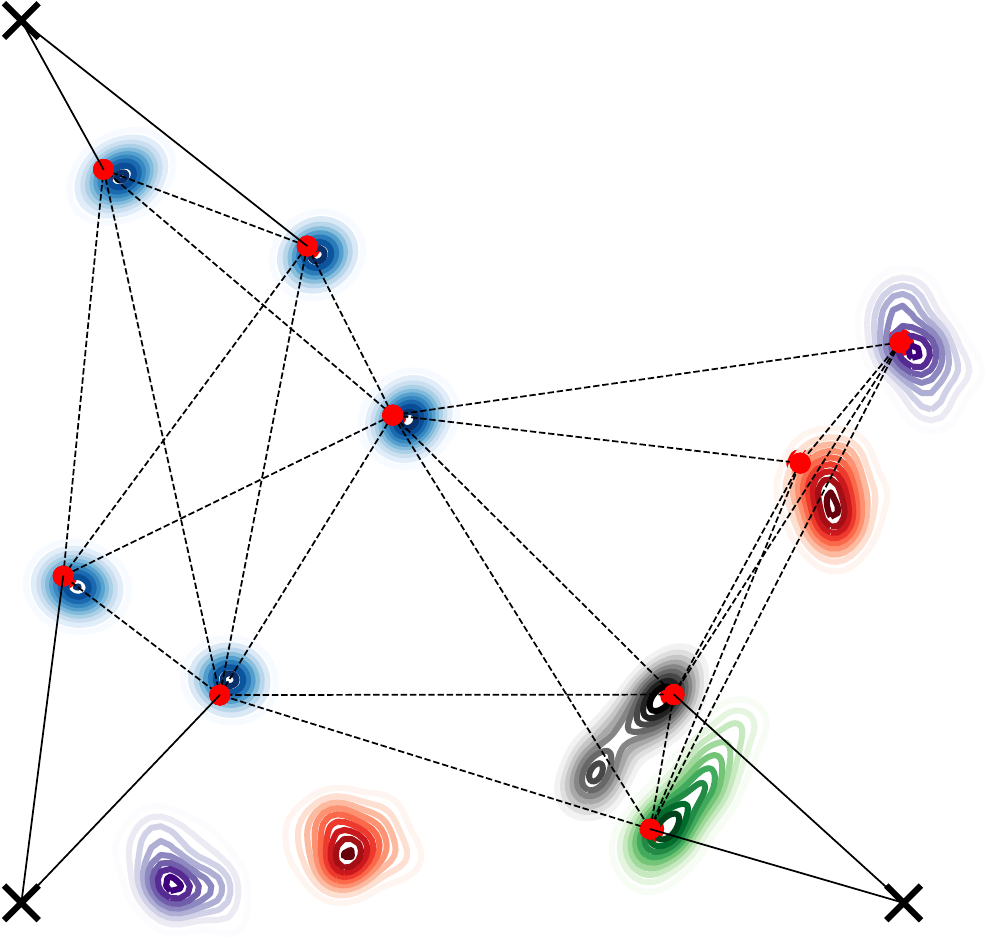} &
~~\includegraphics[height=0.18\textwidth]{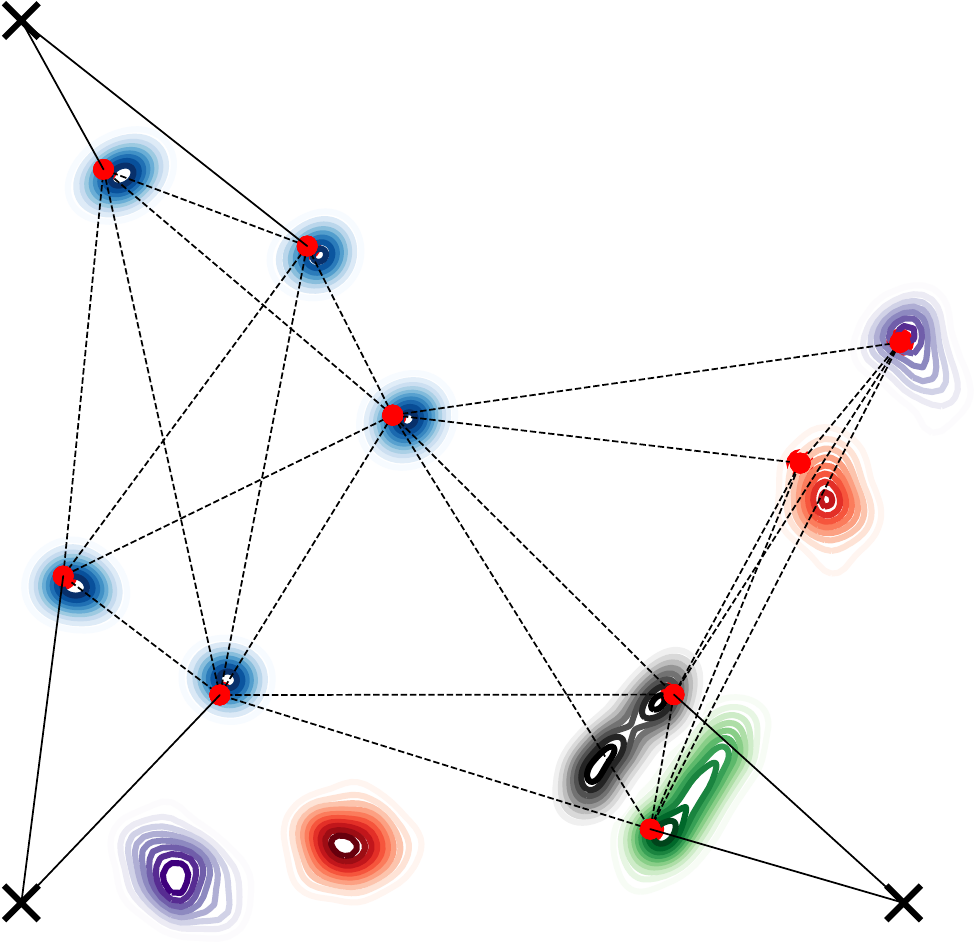}\\
\small  \textit{NUTS} & \small  \textit{SVGD(graphical)} & \small  \textit{SVGD(vanilla)}\\
\includegraphics[height=0.18\textwidth]{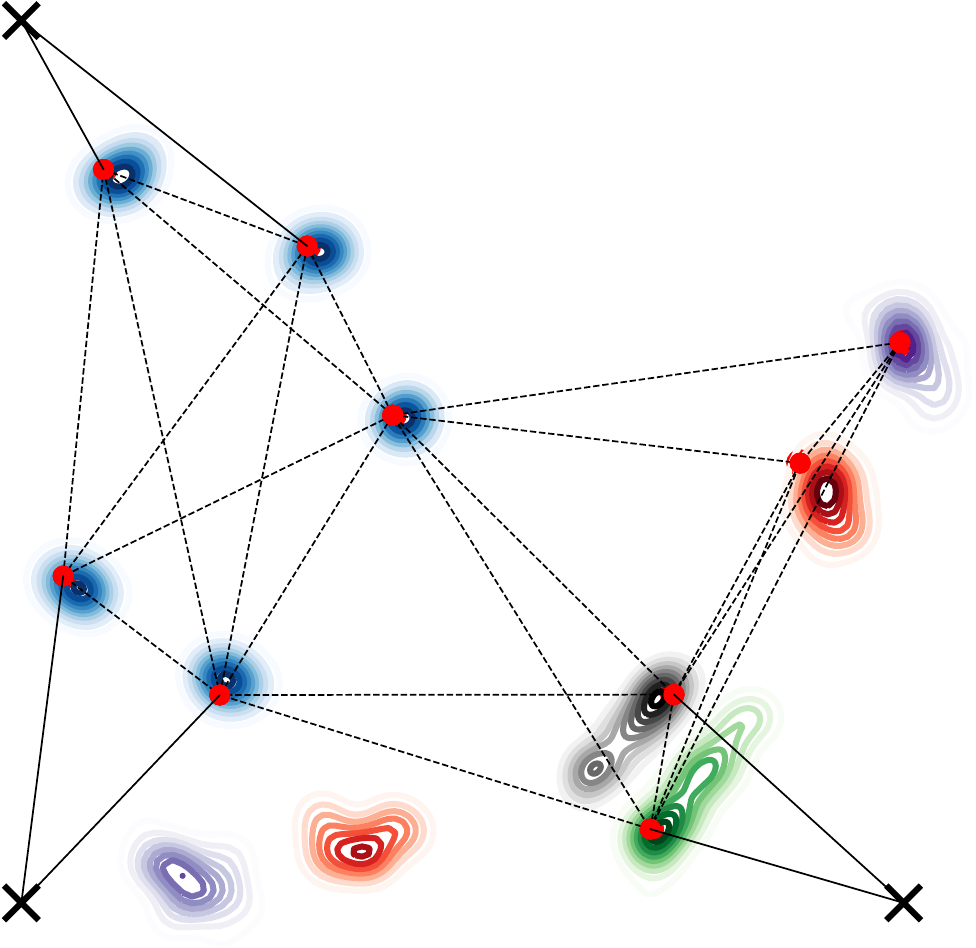} &
~~\includegraphics[height=0.18\textwidth]{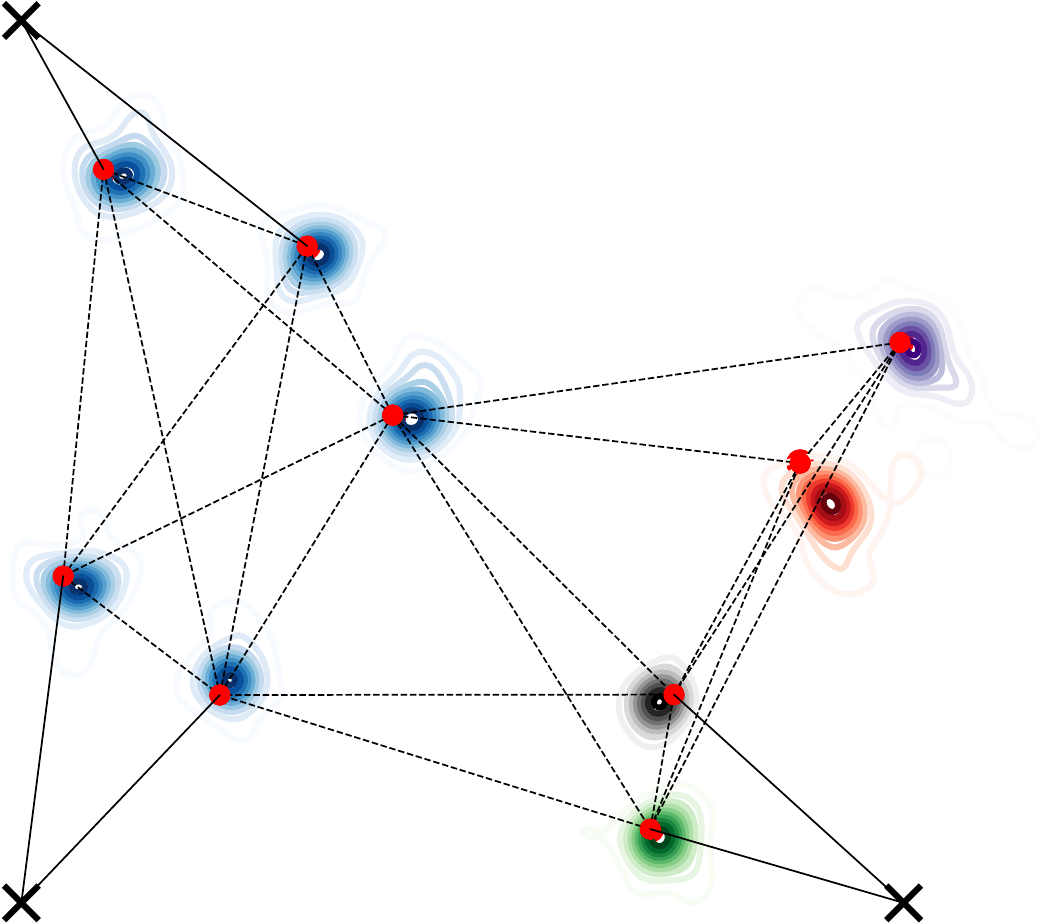} &
~~\includegraphics[height=0.18\textwidth]{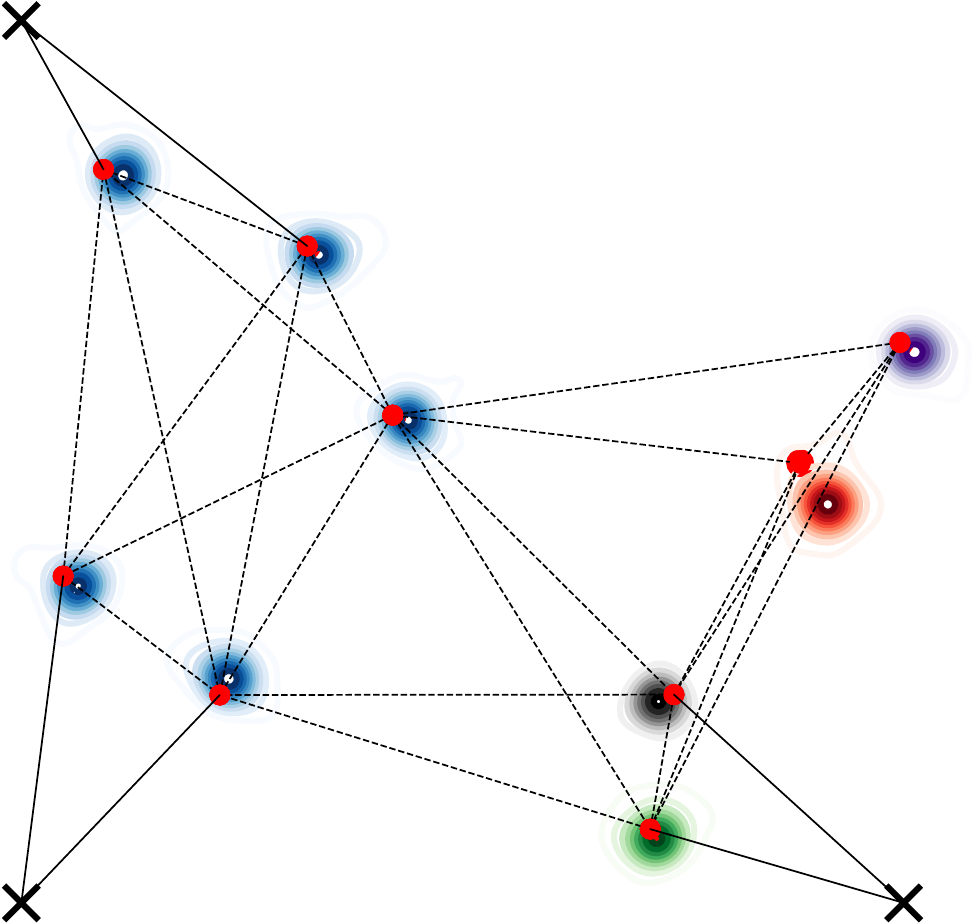} \\
\small \textit{Langevin} & \small  \textit{D-PMP} &\small   \textit{T-PMP} \\
\end{tabular}
\caption{Results on sensor localization in a small network, 
with $m=3$ anchor points (black crosses) and $d=9$ sensors with unknown positions ($d=9$). 
The contours visualize the particles given by different algorithms. 
Because of the lack of anchor point at the upper right corner, 
the posterior locations of node $A$ and $B$ are multi-modal, 
which is recovered by 
\textit{NUTS}, \textit{Langevin}, \textit{SVGD (vanilla)},  and \textit{SVGD(graphical)}, 
but not by  \textit{D-PMP} and \textit{T-PMP}. 
}
\label{fig:sensor_uncertainty}
\end{figure*}

\begin{figure*}[h!]
\centering
\begin{tabular}{ccccc}
\raisebox{1.4em}{\rotatebox{90}{\scriptsize Rooted MSE}}
\includegraphics[width=0.18\textwidth]{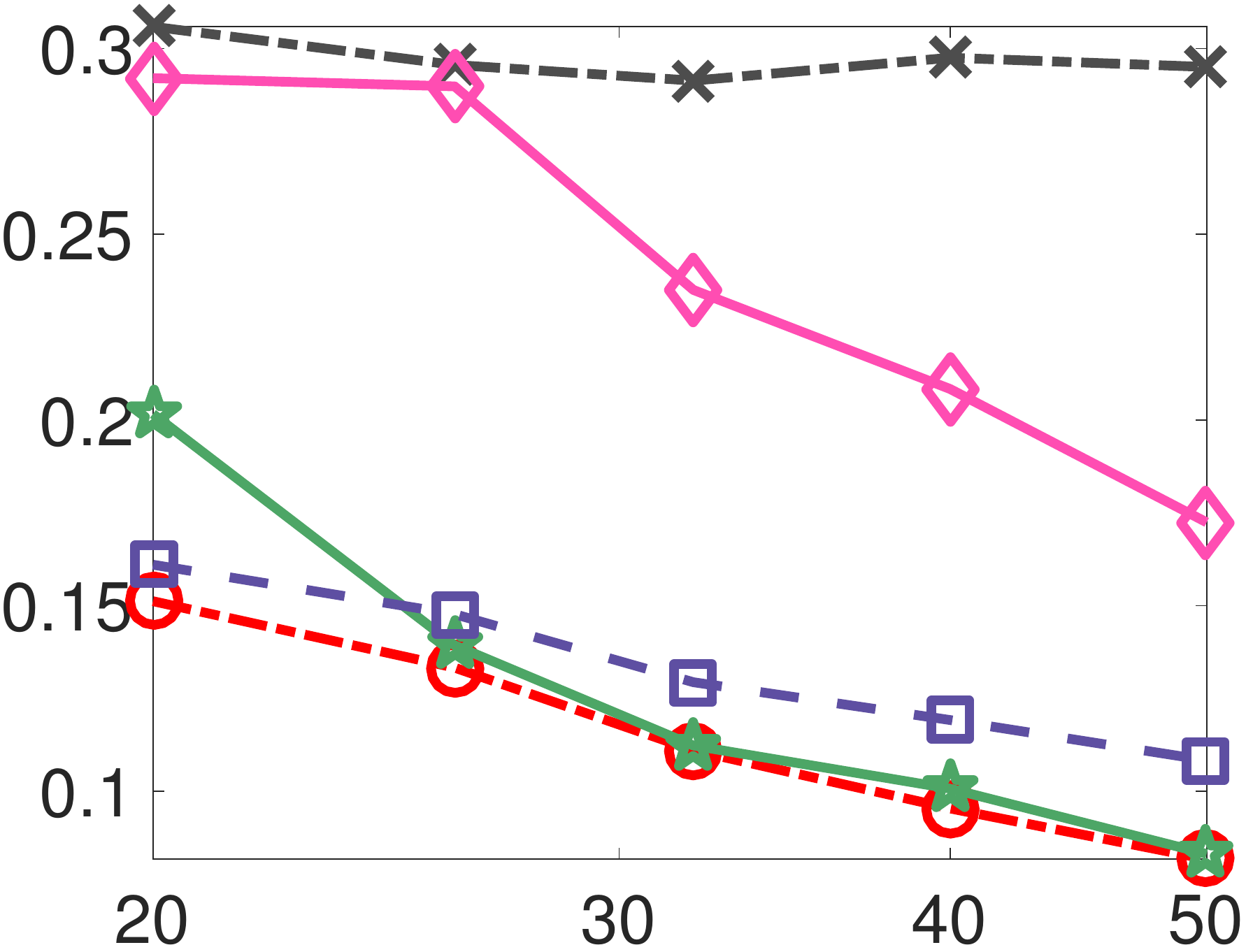} &
\raisebox{1.4em}{\rotatebox{90}{\scriptsize Log10~MSE}}
\includegraphics[width=0.18\textwidth]{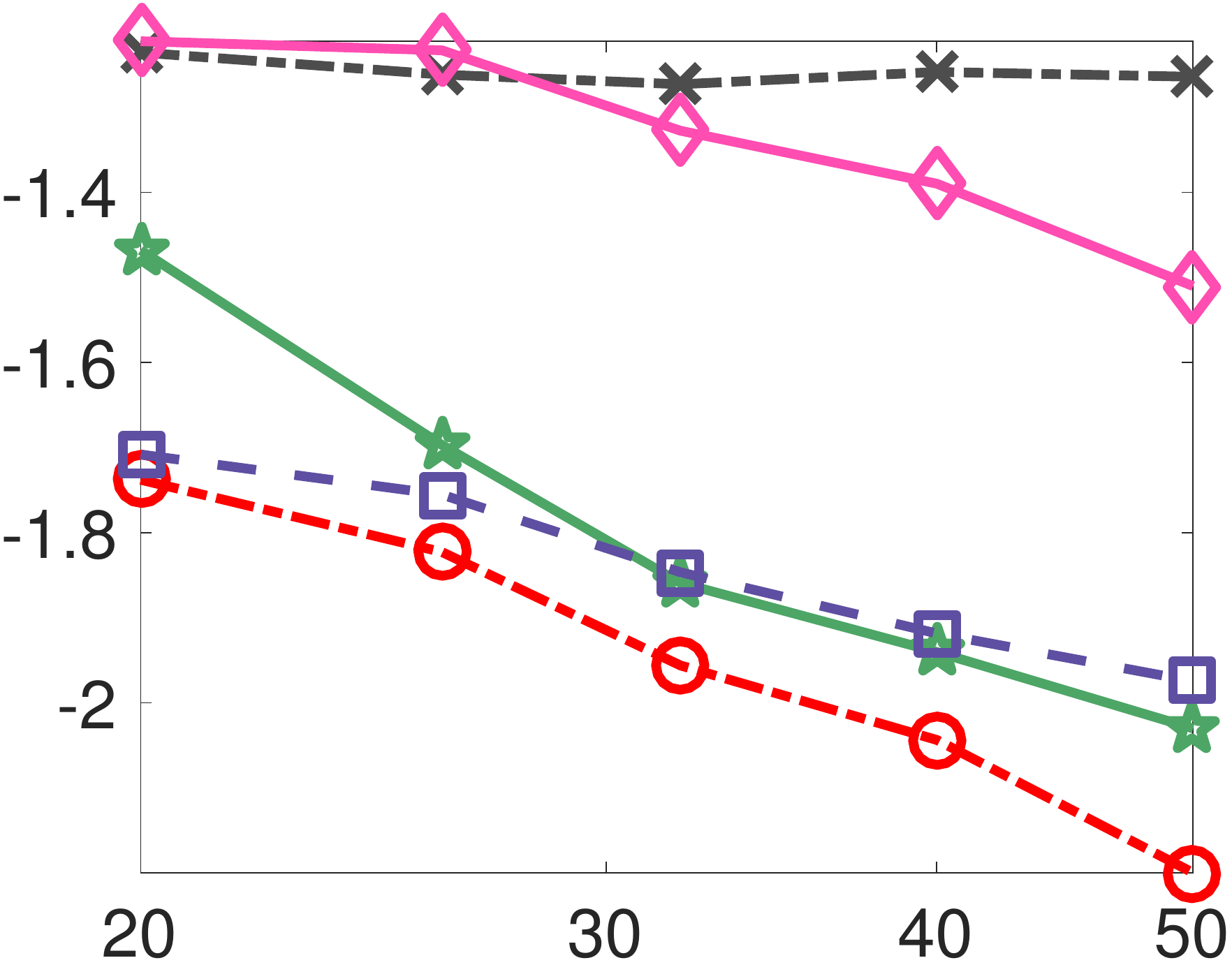} &
\raisebox{1.4em}{\rotatebox{90}{\scriptsize Log10~MSE}}
\includegraphics[width=0.18\textwidth]{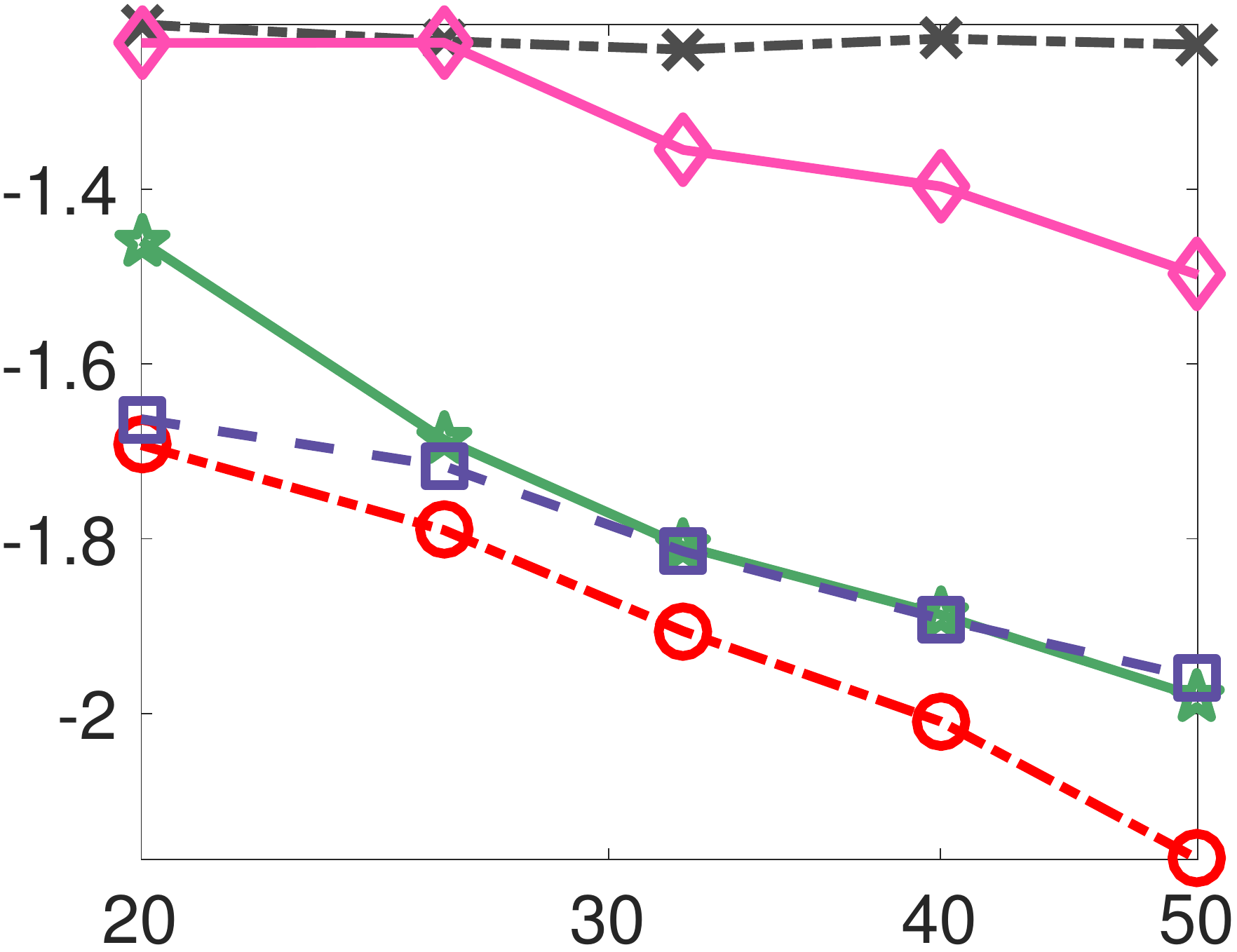} &
\raisebox{1.4em}{\rotatebox{90}{\scriptsize Log10~MMD}}
\includegraphics[width=0.18\textwidth]{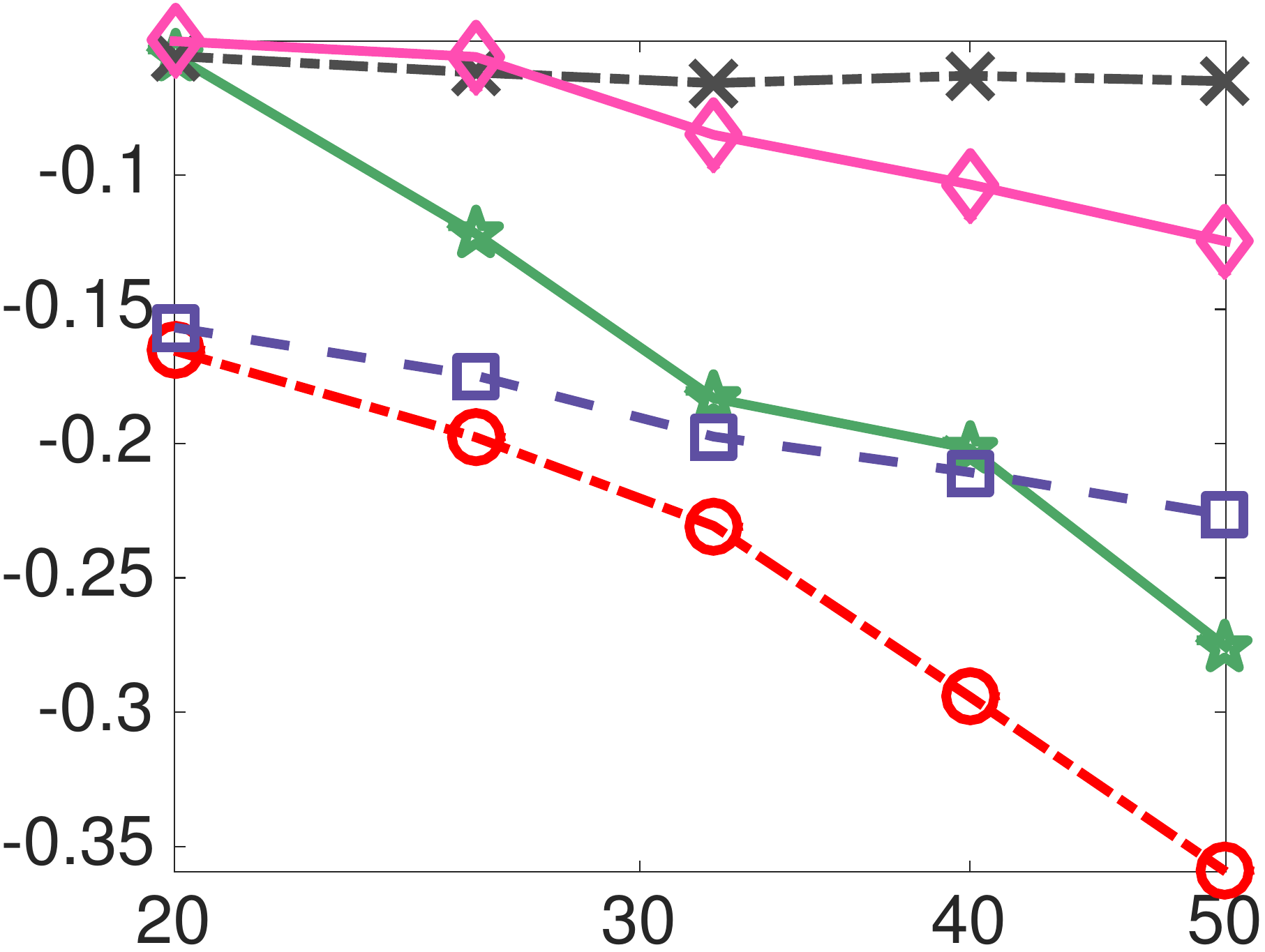} &
\raisebox{3em}{\includegraphics[width=0.1\textwidth]{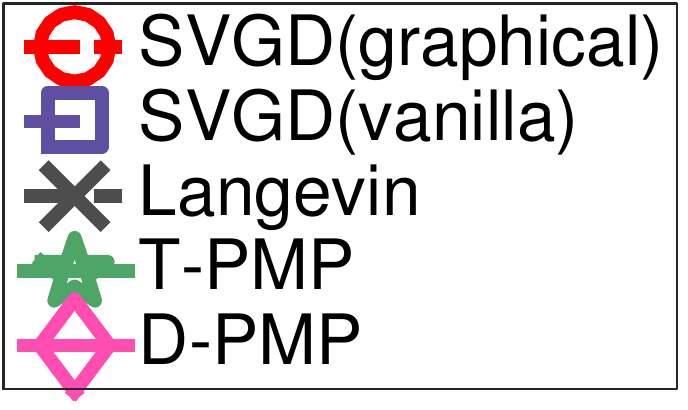}} \\
\scriptsize Number of samples &\scriptsize Number of samples  & \scriptsize Number of samples & \scriptsize Number of samples \\
(a) Localization Error & 
(b) \small Estimating $\E[x_i]$ & (c) \small Estimating $\E[x_i^2]$ & \small (d) MMD vs. $n$ & 
\end{tabular}
\caption{
Results of sensor localization with $m=4$ anchor points and $d = 100$ sensors with unknown locations. 
(a): the rooted mean squared error when estimating the ground truth locations 
using the average of the obtained particles. 
(b)-(d): the estimation accurate in terms of the exact posterior, evaluated against a set of sample obtained by running NUTS for a very large number of steps.  
}
\label{fig:sensor_localization}
\end{figure*}

\section{Experimental Evaluation}





We compare our method with a number of baselines, 
including 
the vanilla SVGD,  particle message passing (PMP), and Langevin dynamics.
Note that Langevin dynamics (without the Metropolis-Hasting (MH) rejection) can also be used in decentralized settings thanks to the factorization form of the gradient $\nabla_x\log p$. 
However, algorithms with MH-rejection, including Metropolis-adjusted Langevin algorithm (MALA) and NUTS \citep{hoffman2014no}, can not be easily performed distributedly because the MH-rejection step requires calculating a global probability ratio.  

We evaluate the results on three sets of experiments, 
including 
a Gaussian MRFs toy example, a sensor localization example, and a crowdsourcing application with real-world datasets. 
We find our method significantly outperforms the baseline methods on  \emph{high dimensional, sparse} graphical models.  

For all our experiments, 
we use Gaussian RBF kernel for both the vanilla and graphical SVGD and choose the bandwidth using the standard median trick. 
Specifically, 
for graphical SVGD, the kernel we use is $k_i(x,x'):= \exp(-{|| x_{\C_i} - x'_{\C_i} ||_2^2}/h_i)$
with bandwidth $h_i = med_i^2$ where $med_i$ is the median of pairwise distances between 
$\{ x_{\C_i}^\ell \}_{\ell=1}^n$ 
for each node $x_i$.
We use AdaGrad \citep{duchi2011adaptive} for step size unless otherwise specified. 

\subsection{Toy example on Gaussian MRFs}

We set our target distribution to be the following
pairwise Gaussian MRF:
$p(x)\propto  \exp [\sum_{i=1}^n ( b_i x_i - \frac{1}{2} A_{ii}x_i^2) - \sum_{(i, j)\in \mathcal{E}} \frac{1}{2}x_i A_{ij}x_j ],$
where $\mathcal E$ is the edge set of a Markov graph. 
The model parameters $(A,b)$ are generated first with $b_i\sim \N(0, 1)$, and $A_{ij} \sim \mathrm{uniform}([-0.1, 0.1])$, followed with $ A \gets (A + A^\top)/2 $ and $A_{ii}\gets  0.1 + \sum_{j\neq i} |A_{ij}|$.

Figure~\ref{fig:gaussian_mrfs}(a)-(c) shows the results when we take $\mathcal E$ to be a 
 4-neighborhood 2D grid of size $10\times 10$, so that the overall variable dimension $d$ equals 100. 
We compare our graphical SVGD (denoted by \textit{SVGD (graphical)}) with a number of baselines, including the typical SVGD (denoted by \textit{SVGD (vanilla)}), exact Monte Carlo sampling, and Langevin dynamics. 
The results are evaluated in three different metrics: 

i) Figure~\ref{fig:gaussian_mrfs}(a) shows the MSE for estimating the mean $\E[x_i]$ of each node $i$, averaged across the dimensions. We see both the graphical and vanilla SVGD perform exceptionally well in estimating the means, which does not seem to suffer from the curse of dimensionality. Graphical SVGD does not show an advantage over vanilla SVGD for the mean estimation. 

ii) Figure~\ref{fig:gaussian_mrfs}(b) shows the MSE for estimating the second order moment $\E[x^2_i]$ of each node $i$, again averaged across the dimensions. In this case, graphical SVGD significantly outperforms vanilla SVGD. 

iii) Figure~\ref{fig:gaussian_mrfs}(c) shows the maximum mean discrepancy (MMD) \citep{gretton2012kernel} between the particles $\{x^\ell\}$ returned by different algorithms and the true distribution $p$, approximated by drawing a large sample from $p$. 
The kernel in MMD is taken to be the RBF kernel, with the bandwidth picked using the median trick. 
We find that vanilla SVGD does poorly in terms of MMD, while graphical SVGD gives the best MMD among all the methods. This is surprisingly interesting because graphical SVGD does not guarantee to match the joint distribution in theory. 


\paragraph{Effects of Graph Sparsity}  
Figure~\ref{fig:gaussian_mrfs}(d) shows the results of graphical and vanilla SVGD as we add more edges into the graph. 
The graphs are constructed by putting $10\times 10$ points uniformly on a 2D grid, and connecting all the points with distance no larger than $r$, with $r$ varying in $[1,\ldots, 14]$.  
The figure shows that the advantage of graphical SVGD compared to vanilla SVGD increases as the sparsity of the graph increases. 

\paragraph{What if we apply local kernels on dense graphs?}   
Although our method requires that the local kernels are strictly positive definite on the Markov blanket of each node, an interesting question is what would happen if the local kernels are defined only on a subset of the Markov blanket, that is, when $k_i(x, x') = k_i(x_{\mathcal D_i}, x_{\mathcal D_i}')$ where $\mathcal D_i$ is a strict subset of $\C_i$ of $p$. To test this, we take $p$ to be a fully connected Gaussian MRF with $(A,b)$ generated in the same way as above, 
and test a variant of graphical SVGD, called \emph{SVGD (graphical+random)}, which uses local kernels  $k_i(x_{\mathcal D_i}, x_{\mathcal D_i}')$ where the domain 
$\mathcal D_i$ is a neighborhood of size five, consisting of node $x_i$ and four  neighbors randomly selected at each iteration. 
The results are shown in Figure~\ref{fig:gaussian_dense}, where we find that \emph{SVGD (graphical+random)} still bring significant improvement over the vanilla SVGD in this case (see Figure~\ref{fig:gaussian_dense}(b)-(c)).  

In Figure~\ref{fig:gaussian_dense}, we also tested \emph{SVGD (combine)} whose kernel is $k_i(x,x') = \alpha k_i(x_{\mathcal D_i}, x_{\mathcal \D_i}') + (1-\alpha) k(x,x')$, which combines the local kernel $\k_i(x_{\mathcal D_i}, x_{\mathcal D_i}')$ with a global RBF kernel $k(x,x')$ ({we take $\alpha = 0.5$}). 
Compared with \emph{SVGD (graphical+random)}, \emph{SVGD (combine)} has the theoretical advantage that $k_i(x,x')$ is strictly integrally positive definite and hence exactly matches the joint distribution $p$ asymptotically.  
Empirically, we find that the performance of \emph{SVGD (combine)} lies in between \emph{SVGD (graphical+random)} and \emph{SVGD (vanilla)} as shown in Figure~\ref{fig:gaussian_dense}. 

\begin{figure*}[t]
\centering
\begin{tabular}{ccccc}
\raisebox{2.0em}{\rotatebox{90}{\scriptsize MSE}}
\includegraphics[width=0.18\textwidth]{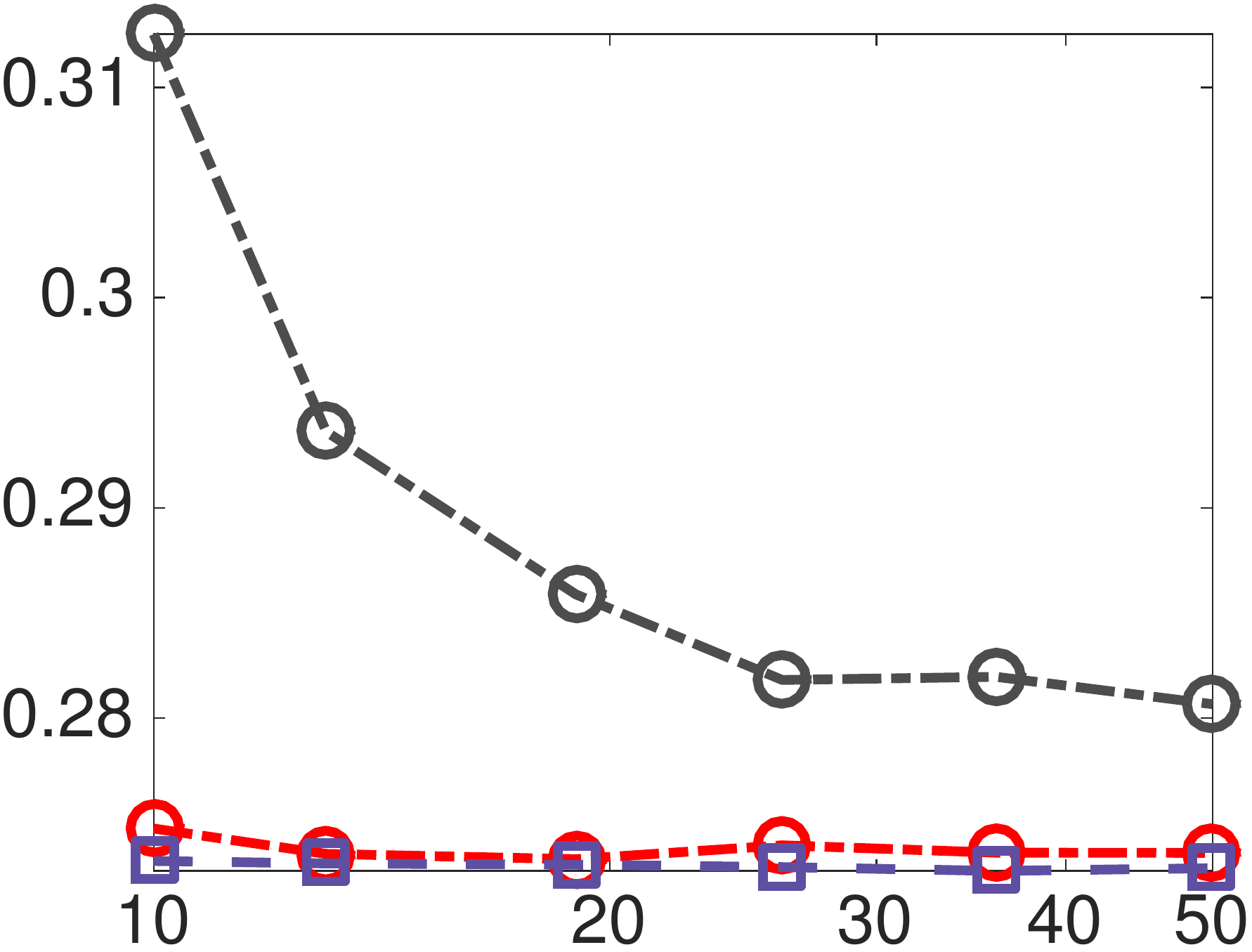}  &
\raisebox{1.4em}{\rotatebox{90}{\scriptsize Log10~MSE}}
\includegraphics[width=0.18\textwidth]{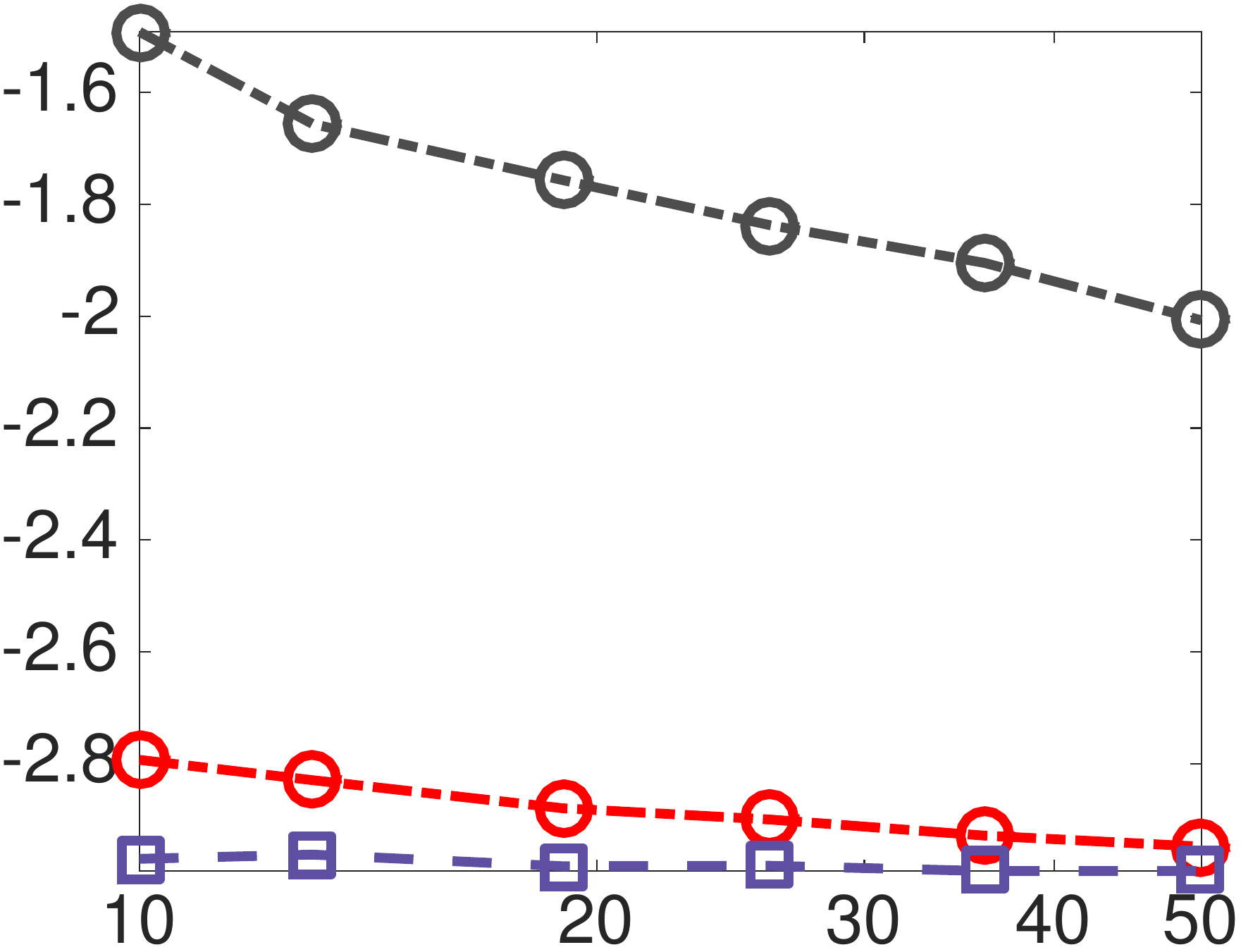}  &
\raisebox{1.4em}{\rotatebox{90}{\scriptsize Log10~MSE}}
\includegraphics[width=0.18\textwidth]{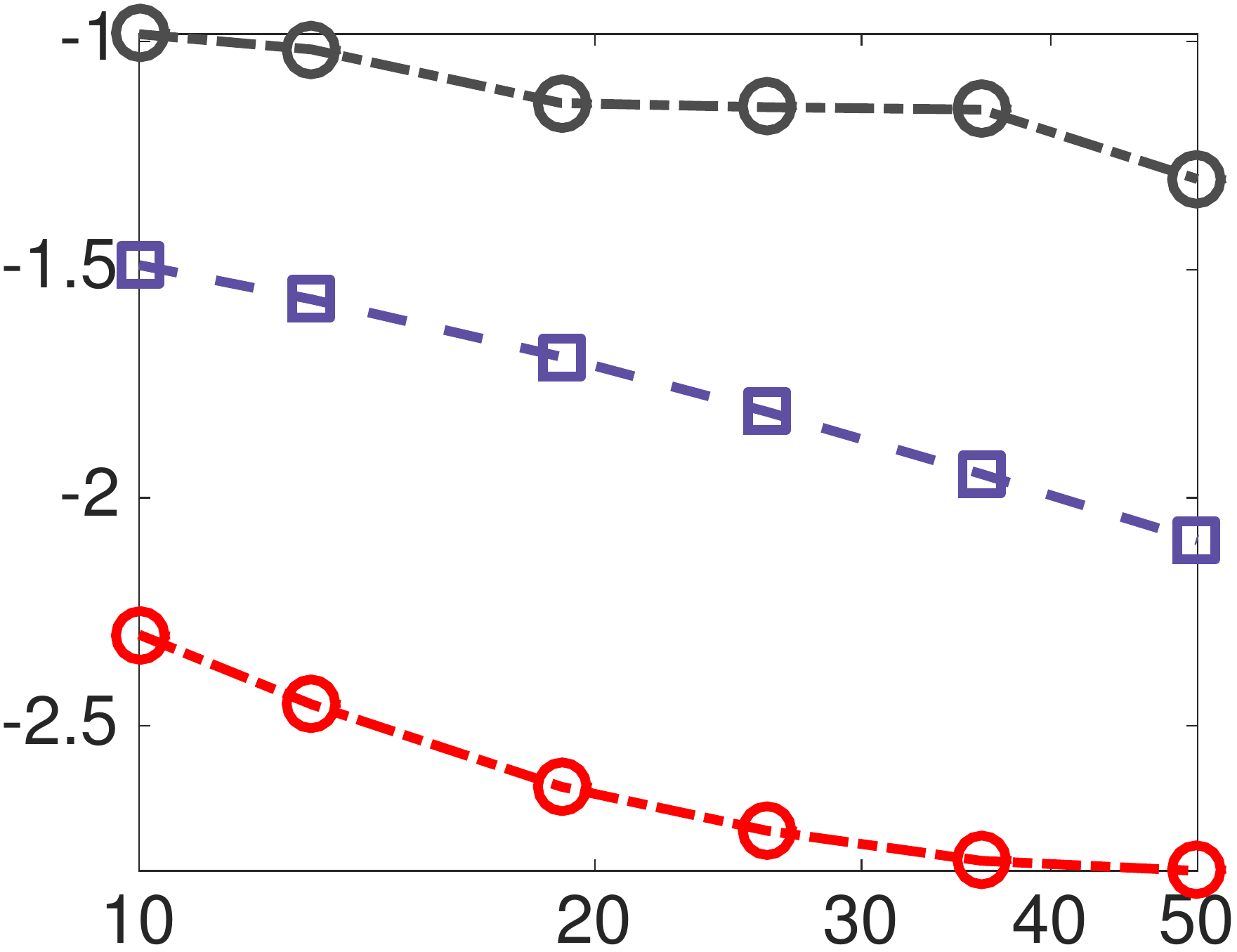}  &
\raisebox{1.4em}{\rotatebox{90}{\scriptsize Log10~MMD}}
\includegraphics[width=0.18\textwidth]{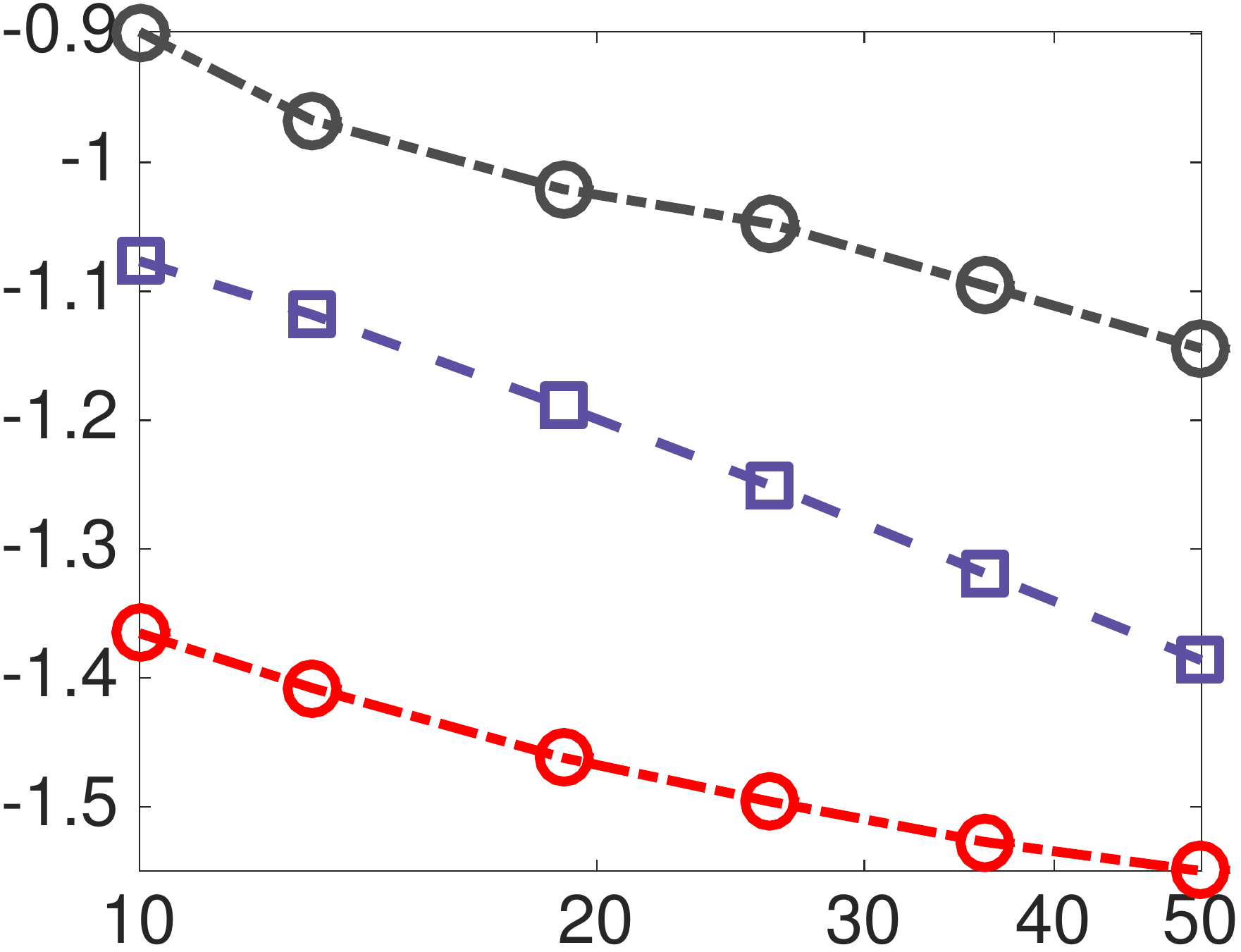}  &
\hspace{-16pt}
\raisebox{3em}{\includegraphics[width=0.1\textwidth]{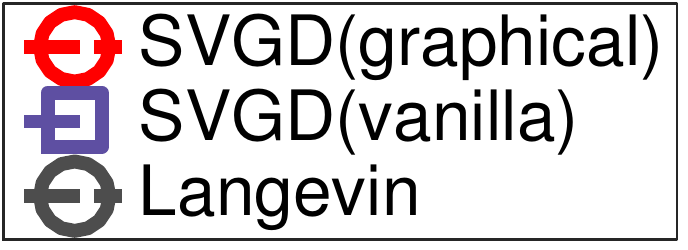}} \\
\scriptsize Number of samples &
\scriptsize Number of samples  & 
\scriptsize Number of samples & 
\scriptsize Number of samples \\
\small (a) MSE w.r.t. true labels &
\small (b) Estimating mean & 
\small (C) Estimating variance &
\small (d) MMD w.r.t. true posterior \\
\end{tabular}
\caption{
Crowdsourcing application on the \emph{PriceUCI} dataset. (a) shows the MSE with respect to the ground truth labels, (b)-(c) the MSE in estimating the posterior mean and variance, and (d) the MMD between the particle approximation and the true posterior. 
}
\label{fig:crowdsourcing_ex}
\end{figure*}
\vspace{-10pt}

\subsection{Sensor Network Localization}

An important task 
in wireless sensor networks 
is to determine the location of each sensor given noisy measurements of pairwise distances \citep[see, e.g.,][]{ihler2005nonparametric}.
We consider a 2D sensor network with nodes consisting of a set $\mathcal S$ of $d$ sensors  placed on unknown locations $
\{x_i\}_{i\in \mathcal S}$, and a set $\mathcal A$ of $m$ anchors with known locations $\{x_i\}_{i\in \mathcal A}$ where $x_i \in \R^{2}$. 
For our experiments, 
we randomly generate the true sensor locations $\{x_i^*\}_{i\in \mathcal S}$ uniformly from interval $[-1,1]^2$.
Assume the sensor-sensor(anchor) distances are measured 
with a Gaussian noise of variance $\sigma$: 
$r_{ij} = ||x_i - x_j||_2 + \sigma \epsilon_{ij}$
where $\epsilon_{ij} \sim \normal (0,1)$ and we set $\sigma = 0.05$. 
Assume only the pairwise distances smaller than 0.5 are measured, and denote the set of measured pairs by $\mathcal E$. The posterior of the  unknown sensor locations is 
$$
p(x_{\mathcal S} | x_{\mathcal A}, r) \propto \prod_{(ij)\in \mathcal E}\exp\left [ -\frac{1}{2\sigma^2} {(||x_i - x_j||_2 - r_{ij})^2}\right ]. 
$$

Particle message passing (PMP) algorithms have been widely used for approximate inference of continuous graphical models, especially for sensor network location \citep{ihler2005nonparametric, ihler2009particle}. 
We compare with two recent versions of PMP methods, including \textit{T-PMP}  \citep{besse2014pmbp} and \textit{D-PMP} \citep{pacheco2014preserving}.  
We use the settings suggested in \citet{pacheco2015proteins}, utilizing a combination of proposals with 75\% neighbor-based proposals and 25\% Gaussian random walk proposals in the augmentation step.
The variance of Gaussian proposals is chosen to be $0.05$,
which is the best setting we found 
on a separate validation dataset simulated with the generative model that we assumed.  
We also select the best learning rate for \textit{Langevin}, \textit{SVGD (vanilla)} and  \textit{SVGD(graphical)} in the aforementioned validation dataset.

Figure \ref{fig:sensor_uncertainty}
reports the contours of 50 particles returned by different approaches on a small sensor network of size $d=9,m=3$, 
which includes three anchor points put on the corners. 
We observe that SVGD and Monte Carlo-type methods tend to capture multiple modes when the location information is ambiguous, while D-PMP and T-PMP obtain more concentrated posteriors. 
For example, the locations of the sensors on $A$ and $B$ are far away from all the three anchor points with known locations and can not be accurately estimated. As a result, the posterior includes two modes $A$, $A'$ and $B$, $B'$, respectively. 
This is correctly identified by both SVGD (graphical), SVGD (vanilla), NUTS and Langevin, 
but not by D-PMP and T-PMP. 

We then demonstrate the effectiveness of our approach on a larger sensor network of size $d=100$. 
We place four anchor sensors at the four corners $(\pm 1, \pm 1)$ ($m=4$), 
In order to evaluate the methods quantitatively, 
Figure~\ref{fig:sensor_localization}(a) shows the mean square error between true locations and the posterior mean estimated by the average of particles. (We find the posterior is essentially unimodal in this case, so posterior mean severs a reasonable estimation).  
Figure~\ref{fig:sensor_localization}(b)-(c) shows the approximate quantity of the posterior distribution, using a set of ground truth samples generated by NUTS \citep{hoffman2014no} with the true sensor positions as the initialization. 
We can see that graphical SVGD tends to outperform all the other baselines. 
It is interesting to see that Langevin dynamics performs significantly worse because it finds difficulty in converging well in this case (even if we searched the best step size extensively). 

\subsection{Application to Crowdsourcing}

Crowdsourcing has been widely used in data-driven applications for collecting large amounts of labeled data. We apply our method to infer unknown continuous quantities from estimations given by the crowd workers.  

Following the setting in \citet{liu2013scoring} and \citet{wang2016efficient}, we assume that there is a set of questions $\{i\}$, each of which is associated with an unknown continuous quantity $x_i$  
that we want to estimate. We assume $x_i$ is generated from a Gaussian prior $x_i \sim \N(0, \sigma_x^2)$. 
Let $\{j\}$ be a set of crowd workers that we  
hire to estimate $\{x_i\}$, and $r_{ij}$ the estimate of $x_i$ given by worker $j$. We assume the label $r_{ij}$ is generated by a bias-variance Gaussian model: 
\begin{align}
r_{ij} = x_i + b_j + \sqrt{\nu_j}\epsilon_{ij}, ~~~ \epsilon_{ij} \sim \N(0,1),
\label{eq:crb_model}
\end{align}
where $b_j$ and $\nu_j$ are the bias and variance of worker $j$, respectively, both of which are unknown with a prior of $b_j \sim \N(0, \sigma_b^2)$ and an inverse Gamma prior $p(\nu_j)=\text{Inv-Gamma}(\alpha, \beta)$ on $\nu_j$. 
We are interested in evaluating the posterior estimation of $\{x_i\}$, which can be done by 
sampling from the joint distribution $p(x,b,\nu \mid r)$. This is a non-Gaussian, highly skewed distribution because it involves the variance parameters $v_j$.


We evaluate our approach on the \textit{PriceUCI} dataset \citep{liu2013scoring}, 
which consists of $80$ household items collected from the Internet, whose prices are estimated by $155$ UCI undergraduate students.
To construct an assignment graph for our experiments, 
we randomly assign 1-5 works to each question, and also ensure each worker is assigned to at least 3 questions. 
Because the bias $b_j$ would not be identifiable without any ground truth, 
we randomly select $10$ questions as control questions with known answers, 
and infer the labels of the remaining $70$ questions. 
The hyper-parameters in the priors are set to be $\sigma_x = \sigma_b = 5$, $\alpha =3, \beta =1$. Results are averaged over 50 random trials. 

We select the best learning rate for \textit{Langenvin}, \textit{SVGD (vanilla)} and our approach
on a separate validation dataset generated with model \eqref{eq:crb_model}.
The inference is applied on model $p(\theta)$ with $\theta=[x, b, \log(\nu)]$,
we clip the value of $\log(\nu)$ to $[-3,3]$ to stablize the training.
For evaluation, we generate a large set of samples for ground truth by running NUTS for a large number steps with the true task labels as initialization. 
As shown in figure \ref{fig:crowdsourcing_ex}, 
our graphical SVGD again outperforms both the typical SVGD and Langevin dynamics.  


%% file: tex/appendix.tex

\appendix
\numberwithin{equation}{section}

\appendix
\numberwithin{equation}{section}

\section{ Proof of Theorem \ref{thm:optimal}}

\begin{proof}
Consider $\vv f = [f_1,\ldots, f_d]^\top \in \H$, where $\H  = \H_1\times\cdots \times \H_d$. 
Using the reproducing property of $\H_i$, we have for any $f_i \in \H_i$ 
$$
\begin{aligned}
&\E_{x\sim q}[\steinp_{x_i} f_i(x)] = \langle f_i , ~ \E_{x \sim q}[\steinp_{x_i} k_i(x, \cdot)] \rangle_{\H_i}. \\
\end{aligned}
$$

Recall that $\phi^*_i(\cdot) =\E_{x \sim q}[\steinp_{x_i} k_i(x,\cdot)]$, and $\ff^* = [\phi^*_1,\ldots, \phi^*_d]^\top$. 
The optimization of the Stein Discrepancy is framed into
$$
\begin{aligned}
\S(q~||~p) & = \max\limits_{\vv f \in \H, ||\vv f||_{\H} \le 1} \E_{x \sim q}[\steinpx^\top \vv f(x) ] \\
& = \max\limits_{\vv f \in \H, ||\vv f||_{\H} \le 1} \sum_{i=1}^d \E_{x \sim q}[\steinp_{x_i} f_i(x) ] \\
& = \max\limits_{\vv f \in \H, ||\vv f||_{\H} \le 1} \sum_{i=1}^d  \langle f_i, \phi^*_i \rangle_{\H_i}\\
& = \max\limits_{\vv f \in \H, ||\vv f||_{\H} \le 1} \langle \vv f, \vv \phi^*  \rangle_{\H}. 
\end{aligned}
$$
This shows that the optimal $\vv f$ should equal $\ff^*/||\ff^*||_\H$,
and $\S(q~||~p) = \langle \vv \ff^*/||\ff^*||_\H, \vv \phi^*  \rangle_{\H} = ||\ff^*||_\H$.  
\end{proof}

\section{ Proof of Theorem \ref{lem:graphical_svgd_update}}
\begin{proof}
Plugging the optimal solution in Theorem \ref{thm:optimal} into the definition of Stein discrepancy \eqref{equ:ksd_prob}, we get 
$$
\begin{aligned}
\S(q~||~p) 
& = \frac{1}{||\ff^*||_\H} \E_{x \sim q}[\steinpx^\top \ff^*(x)]   \\
& = \frac{1}{||\ff^*||_\H}  \sum_{i=1}^d \E_{x \sim q}[\steinp_{x_i}\phi_i^*(x)]  \\
& = \frac{1}{||\ff^*||_\H}  \sum_{i=1}^d \E_{x,x'\sim q}[\steinp_{x_i} \steinp_{x_i'} k_i(x,x')].\\ 
\end{aligned}
$$
On other hand, because $\S(q~||~p) = ||\ff^*||_\H$, we have 
\begin{align}\label{tmdfdppp}
\S(q~||~p)^2  = \sum_{i=1}^d \E_{x,x'\sim q}[\steinp_{x_i} \steinp_{x_i'} k_i(x,x')]. 
\end{align}

To prove \eqref{equ:deltaq}, note that 
\begin{align*}
    & \E_{\vx \sim q}[\steinp_{x_i} f (\vx)] \\
    &  = \E_{\vx\sim q}[\steinp_{x_i} f ]  - \E_{q}[\mathcal Q_{x_i} f ]   \\
    & = \E_{\vx \sim q} [(\nabla_{x_i} \log p(\vx) - \nabla_{x_i}\log q(\vx) ) f(x)]  \\
    & = \E_{\vx \sim q} [(\nabla_{x_i} \log p(x_i | \vx_{\neg i}) - \nabla_{x_i}\log q(x_i | \vx_{\neg i}) ) f(x)] \\
    & = \E_{\vx \sim q} [\delta_i(\vx) f(\vx)]. 
\end{align*}
Applying this equation twice to \eqref{tmdfdppp} gives 
\begin{align}\label{gkd}
\S(q~||~p)^2 = \sum_{i=1}^d \E_{x,x'\sim q} [ \delta_i(x) k_i(x,x') \delta_i(x')]. 
\end{align}

By \eqref{gkd} and the definition of strictly integrally positive definite kernels, 
we can see that $\S(q~||~p)=0$ implies $\delta_i(x) = 0$, $\forall i\in [d]$, if $k_i(x,x')$ is strictly integrally positive definite for each $i$. 
Note that $\delta_i(x) =0$ means $p$ and $q$ matches the conditional probabilities: 
\begin{align} 
p(x_i | x_{\neg i}) = q(x_i | x_{\neg i}), ~~~ \forall i \in [d].  
\end{align}
For positive densities, this implies that $p(x) =q(x)$ (see e.g., \citet{brook1964distinction,besag1974spatial}). 
\end{proof}

\section{ Proof of Theorem \ref{thm:graphical_svgd_update}}
\begin{proof}
For a graphical model $p(x)$ with Markov blanket $\mathcal N_i$ for node $i$, we have
$$\nabla_{x_i} \log p(x_i | x_{\neg i}) = \nabla_{x_i} \log p(x_i | x_{\N_i})~~~~~\forall i\in[d].$$
Moreover, by Stein's identity  on $q$, we have 
$$\E_{x \sim q}[\nabla_{x_i}\log q(x_i \mid x_{\N_i}) f(x) + \nabla_{x_i}f(x)] = 0,~~~\forall i\in [d].$$ 
With a similar argument as the proof of Theorem  \ref{lem:graphical_svgd_update}, we get 
$$
\begin{aligned}
& \E_{x \sim q}[\steinp_{x_i}f(x)] \\
& = \E_{x \sim q}[  (\nabla{x_i} \log p(x_i | x_{\N_i}) - \nabla{x_i} \log q(x_i | x_{\N_i})) f(x)] \\
& = \E_{x \sim q}[ \delta_i(x_{\C_i}) f(x) ], 
\end{aligned}
$$
where $\delta_i(x_{\C_i}) = \nabla_{x_i} \log q(x_i | x_{\N_i}) - \nabla_{x_i} \log p(x_i | x_{\N_i})$.

Applying this equation twice to \eqref{tmdfdppp} gives 
$$\S(q~||~p)^2 = \sum_{i=1}^d 
\E_{x,x'\sim q} [ \delta_i(x_{\C_i}) k_i(x, x') \delta_i(x_{\C_i}')].$$
Therefore, if $k_i(x,x')$ is strictly integrally positive definite on $x_{\C_i}$, 
Stein discrepancy $\S(q~||~p)=0$ if and only if $q(x_i | x_{\N_i})  = p(x_i | x_{\N_i})$. 

\end{proof}